\documentclass[lettersize,journal]{IEEEtran}
\usepackage{amsmath,amsfonts}
\usepackage{algorithmic}
\usepackage{algorithm}
\usepackage{array}
\usepackage{textcomp}
\usepackage{diagbox}
\usepackage{url}
\usepackage{verbatim}
\usepackage{graphicx}
\usepackage{amsmath,amssymb}
\usepackage{psfrag}
\usepackage{epsfig}
\usepackage{cite}
\usepackage{ulem}
\usepackage{graphicx}
\usepackage{makecell}
\usepackage{multirow}
\usepackage{threeparttable}
\usepackage{booktabs}
\usepackage{mathrsfs}
\usepackage{amsfonts}
\usepackage{amsmath}
\usepackage{psfrag}
\usepackage{graphicx}
\usepackage{amssymb}
\usepackage{algorithmic}
\usepackage{algorithm}
\usepackage{color}
\usepackage{overpic}
\usepackage{cases}
\usepackage{listings}
\usepackage {setspace}
\usepackage{cite}
\usepackage{pifont}
\usepackage{amsmath,amssymb,amsfonts}
\usepackage{algorithmic}
\usepackage{graphicx}
\usepackage{textcomp}
\usepackage{amsmath,amssymb}
\usepackage{enumerate}
\usepackage{psfrag}
\usepackage{ragged2e}
\usepackage{url}
\usepackage{epsfig}
\usepackage{color}
\usepackage{graphics}
\usepackage{tabularx}
\usepackage{booktabs}
\usepackage[T1]{fontenc}
\usepackage{fancybox}
\usepackage{multirow}
\usepackage{threeparttable}
\usepackage{booktabs}
\usepackage{mathrsfs}
\usepackage{bm}
\usepackage{algorithm}
\usepackage{algorithmic}
\usepackage{ulem}
\usepackage{subfigure}
\usepackage{media9}
\usepackage[labelfont=bf,format=plain,justification=raggedright,singlelinecheck=false]{caption}
\usepackage[colorlinks=true,linkcolor=black,urlcolor=black,citecolor=black]{hyperref}
\usepackage{nomencl}
\makenomenclature
\begin{document}

\title{Collision Risk Quantification and Conflict Resolution in Trajectory Tracking for Acceleration-Actuated Multi-Robot Systems}

 \author{Xiaoxiao~Li,
             Zhirui~Sun,
             Mansha~Zheng,
             Hongpeng~Wang,
             Shuai~Li,~\IEEEmembership{Senior~Member,~IEEE,}
             Jiankun~Wang,~\IEEEmembership{Senior~Member,~IEEE}
\thanks{This work was supported in part by the Guangdong Provincial Key Laboratory of Novel Security Intelligence Technologies under Grant 2022B1212010005, in part by the Shenzhen Basic Research Project (Natural Science Foundation) under Grant JCYJ20210324132212030. (Corresponding author: Hongpeng Wang)}
\thanks{X. Li, M. Zheng and H. Wang are with the Harbin Institute of Technology Shenzhen, P.R. China. (e-mails: lxx@stu.hit.edu.cn; mszheng@stu.hit.edu.cn; wanghp@hit.edu.cn).}
\thanks{H. Wang is also with the Guangdong Provincial Key Laboratory of Novel Security Intelligence Technologies.}
\thanks{Z. Sun and J. Wang are with the Department of Electronic and Electrical Engineering, Southern University of Science and Technology, Shenzhen 518055, China (e-mail: sunzr2023@mail.sustech.edu.cn; wangjk@sustech.edu.cn).}
\thanks{S. Li is with the Faculty of Information Technology and Electrical Engineering, University of Oulu, Finland, and also with the VTT-Technical Research Centre of Finland, 90590 Oulu, Finland (e-mails: shuai.li@oulu.fi).}}


\markboth{IEEE Transactions on Industrial Electronics}%
{Shell \MakeLowercase{\textit{et al.}}: Collision Risk Quantification and Conflict Resolution in TT for Multi-Robot Systems}


\maketitle

\begin{abstract}
One of the pivotal challenges in a multi-robot system is how to give attention to accuracy and efficiency while ensuring safety.
Prior arts cannot strictly guarantee collision-free for an arbitrarily large number of robots or the results are considerably conservative. Smoothness of the avoidance trajectory also needs to be further optimized.
This paper proposes an acceleration-actuated simultaneous obstacle avoidance and trajectory tracking method for arbitrarily large teams of robots, that provides a non-conservative collision avoidance strategy and gives approaches for deadlock avoidance. We propose two ways of deadlock resolution, one involves incorporating an auxiliary velocity vector into the error function of the trajectory tracking module, which is proven to have no influence on global convergence of the tracking error. Furthermore, unlike the traditional methods that they address conflicts after a deadlock occurs, our decision-making mechanism avoids the near-zero velocity, which is much more safer and efficient in crowed environments.
Extensive comparison show that the proposed method is superior to the existing studies when deployed in a large-scale robot system, with minimal invasiveness.
\end{abstract}

\begin{IEEEkeywords}
Obstacle avoidance, deadlock resolution, multi-robot systems, optimization.
\end{IEEEkeywords}

\mbox{}
\printnomenclature

\section{Introduction}
\subsection{Motivation}
\IEEEPARstart{C}{ollision} avoidance (CA) is a perennial topic for multiple wheeled mobile robots (MWMRs) systems. Although numerous CA methods have been proposed, such as a series of variants \cite{Cooperative2018TRO,zhang2020cooperative,LRA2022} that build on the concept of the velocity obstacle (VO), time-to-collision-based methods \cite{shahriari2021toward}, dynamic vector field based methods \cite{he2024simultaneous}, \emph{etc}, how to give attention to both accuracy and efficiency while ensuring that each wheeled mobile robot (WMR) does not collide with other wheeled mobile robots (WMRs) during the trajectory tracking (TT) still is one of the fundamental challenges in a MWMRs system.
As the number of WMRs increases, WMRs spend most of the time in avoiding the impending collisions and, as a result, they overly deviate from their preferred trajectories.

A remedy to this problem requires avoiding the collision and deadlock, while be as close as possible to the preferred trajectory. It is necessary to design a controller with minimally invasive CA.
Compared to VO with enlarged conservative bounding volumes and learning-based CA methods with probabilistic safety guarantee, control barrier function (CBF) provides the minimum modification necessary to formally guarantee safety in the context of quadratic programming (QP), and strict safety guarantee for safety-critical systems.
Therefore, our method builds on the context of the CBF.
Safety barrier certificate (SBC) method is proposed in \cite{IFAC2015} which extends the CBF to the MWMRs system by incorporating all pairwise collision-free constraints into an admissible control space. The SBC is further applied to heterogeneous swarm with different maximum accelerations in \cite{ACC2016}, and distributes CA responsibilities for WMRs based on their maximum acceleration in \cite{Safety2017TRO}. However, \cite{IFAC2015,ACC2016,Safety2017TRO} omits the WMRs’s kinematics constraint. More importantly, a main drawback of the SBC is that it cannot strictly guarantee collision-free for all WMRs in a crowed environment because of the introduction of the braking force. Most importantly, the SBC solves the synthesized QP equation using the MATLAB \emph{quadprog} solver which may find a solution slowly and be sensitive to initial values, and not suitable to the real-time application.

The follow-up CBF-related research has increasingly focused on the design of robust safety controllers disturbed by both measurement uncertainty and model uncertainty, such as \cite{luo2020multi,xiao2023barriernet,marvi2021safe}, to name a few.
As the number of WMRs increases, deadlock conflict exacerbates. In a deadlock
state, despite safety, WMRs come to a standstill before reaching their goals.
In \cite{SMC2017}, token-based random allocation method is used to determine WMRs's actions. However, WMRs cannot change their paths. Both collision and deadlock are avoided by repeatedly stopping and resuming WMRs.  Interaction between WMRs can be modeled using a game formulation \cite{TCST2021}, or learning-based methods \cite{LRA2022} are used to enhance both CA and deadlock prevention ability of the aforementioned traditional methods. However, design challenge of game strategy, communication requirement and poor generalization hinder the real-time application of these two methods from simulation to real-world in a large-scale robot scenario. Rule-based \cite{Safety2017TRO,he2024simultaneous} and priority-based \cite{priority2024} planning are two common path coordination methods.
In our work, we follow the idea of \cite{Safety2017TRO,he2024simultaneous}. In contrast to previous efforts, we propose two ways of deadlock resolution, one involves incorporating an auxiliary vector into the error function of the TT module, which is proven to guarantee global convergence of the tracking error, rather than boundedness. The another is to modify the preferred velocity. Comparison shows these two ways outperform those in \cite{Safety2017TRO} and \cite{he2024simultaneous}. Furthermore, unlike the traditional methods that they address conflicts after a deadlock occurs, our decision-making mechanism avoids the near-zero velocity, which is much more safer and efficient in crowed environments.
\vspace{-13pt}
\subsection{Related Work}
\begin{figure*}
  \centering
  \includegraphics[width=7.1in]{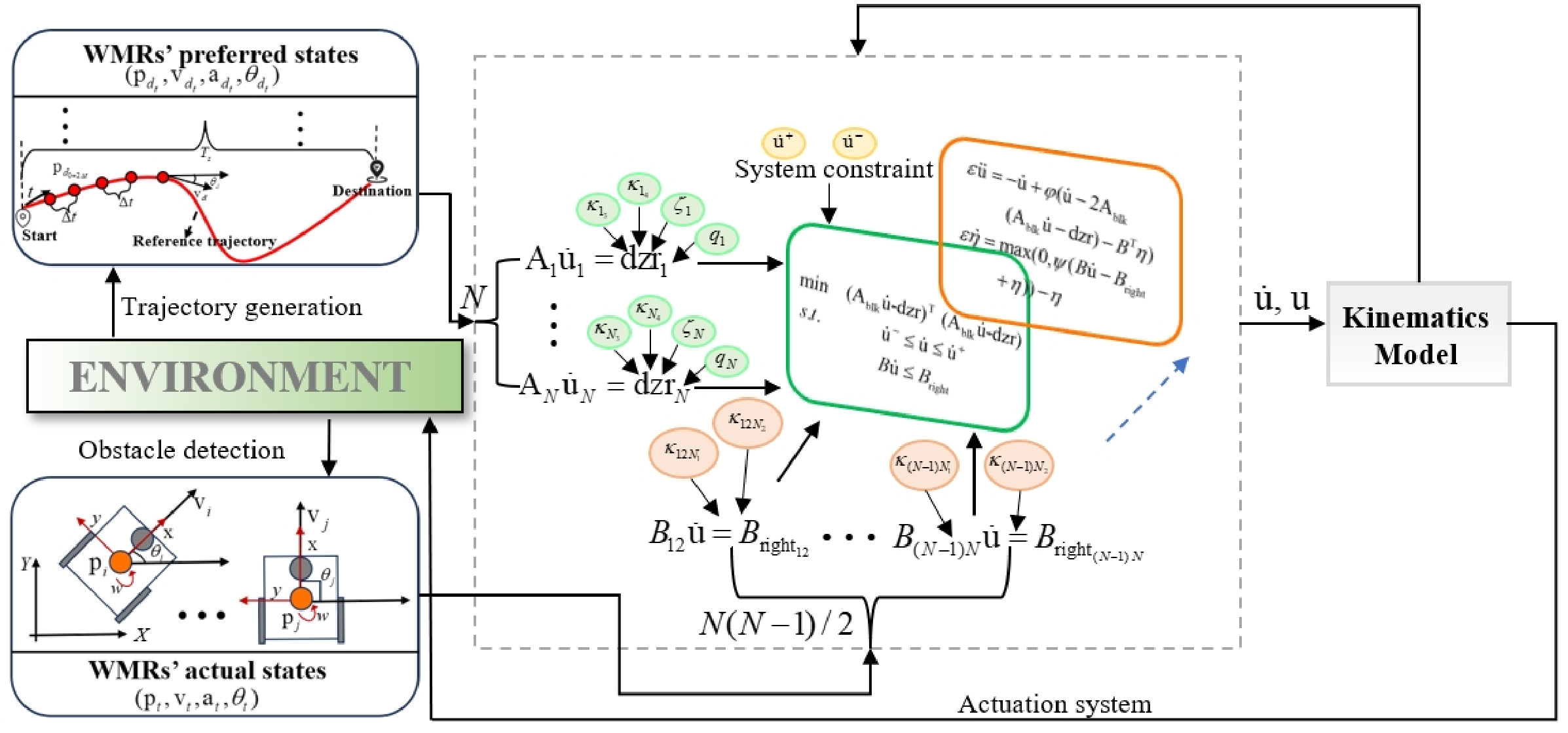}
  \caption{\justifying Schematic of the AA-SOATT-based control method. WMRs obtain states information of the neighboring WMRs and obstacles relying on the onboard sensors. These data, along with preferred states and physical constraints, are fed into the system, and as a result, a QP formulation is synthetized. By solving this equation recursively, the optimal control inputs are derived and then employed in conjunction with the WMR's kinematics model to actuate WMRs.}\label{h}
\end{figure*}

The CBF is first conceptualized in \cite{wieland2007constructive}, which is use to map an inequality constraint defined over system states onto a constraint on the control input. \cite{ames2014control} establishes the unification of control Lyapunov functions and CBF  through a QP for simultaneous achievement of multi-objectives. The CBF is refined as the zeroing CBF (ZCBF) and reciprocal CBF in \cite{ames2016control} to adapt to different control requirements and system characteristics.
The SBC method is proposed in \cite{IFAC2015} and is further applied to heterogeneous swarm with different maximum accelerations in \cite{ACC2016}, and distributes CA responsibilities for WMRs based on their maximum acceleration in \cite{Safety2017TRO}.
The SBC method lays the foundation for ensuing research on CBFs-based MWMRs motion coordination. However, the follow-up research has focused more on the design of robust safety controllers. \cite{marvi2022robust} incorporates information gap theory into the SBC, proposing a robust safety controller for multi-agent systems with different measurement accuracies. \cite{capelli2021decentralized} considers the safety control problem for MWMRs under communication delays. To against model uncertainty and measurement uncertainty disturbances, Luo \emph{et al.} propose probabilistic SBC \cite{luo2020multi}. Although it enhances safety of SBC in the presence of uncertainty, the probabilistic constraints generate a more enlarged admissible control space and may make WMRs' behavior more aggressive, potentially exacerbating deadlock. The combination of CBF with learning-based methods has also made progress, primarily divided into two branches. One is to use learning as a controller. \cite{xiao2023barriernet} proposes differentiable CBF to ensure the safety of control models learned for autonomous driving tasks. \cite{marvi2021safe} introduces a damping coefficient into CBF to regulate the trade-off between safety and optimality, using a non-policy reinforcement learning algorithm to find a safety-optimal policy without complete knowledge of system dynamics. The another is that learning safety constraints from data. By collecting historical data, safety boundaries under specific conditions can be learned, helping to define CBF more suited to particular environments. \cite{wang2018safe} learns models of quadrotors operating in partially unknown environments. \cite{lindemann2021learning} establishes sufficient conditions regarding data to learn hybrid CBF for systems with both continuous and discrete states and transitions. We have not elaborated on all theoretical research and applications related to CBF. \cite{li2023survey} and \cite{anand2021safe} have systematically reviewed CBF-related research from perspectives of nonlinear control systems and reinforcement learning.
These proposals are aimed at addressing the bottlenecks and challenges in ensuring security using CBF.
Our work is most closely related to the work on SBC, aiming at synthesizing an acceleration-actuated simultaneous obstacle avoidance and trajectory tracking (AA-SOATT) framework of minimally CA invasiveness for arbitrarily large teams of WMRs. This is a feature not possible with the SBC.
Moreover, the proposed method does not communicate with each other and only relies on the WMR’s states, and we present in an analytical expression, so
that it can be straightforwardly computed in real time as long as the WMR’s states are obtained. Schematic of the AA-SOATT-based control method is given in Fig. \ref{h}.
\vspace{-10pt}
\subsection{Contributions}
The main contributions of this paper include:
\begin{itemize}
\item[1)]The AA-SOATT methods is proposed for arbitrarily large teams of WMRs, that provide a nonconservative CA strategy and
give approaches for deadlock avoidance.
\item[2)]We propose two ways of deadlock resolution, one involves incorporating an auxiliary velocity vector into the
error function of the TT module, which is proven to guarantee global convergence of the tracking error.
\item[3)]The designed control law can globally converge to a unique equilibrium point and is globally asymptotically stable in the Lyapunov sense.
\end{itemize}
\vspace{-10pt}
\subsection{Organization}
 Section \ref{sec.2} gives the system model and problem statement, and then briefly recalls the CBF, as well as introduces some CA functions for the ensuing comparison.
 The AA-SOATT methods are proposed in Section \ref{sec.3}. Subsequently, the design of the control law and theoretical analyses are given in Section \ref{sec.4}. Results and analyses are illustrated in Section \ref{sec.5}. Section \ref{sec.6} concludes this paper.
\section{Preliminaries and Problem Formulation}\label{sec.2}
\subsection{System Model}
Utilizing a double integrator and feedback linearlization technology \cite{Li2022SOATT}, the motion of the robot $i$ can be described as:
 \begin{equation}\label{eqn.kinematic}
\left[\begin{array}{c}
\dot{\mathbf{p}}_i \\
\dot{\mathbf{v}}_i
\end{array}\right]=\left[\begin{array}{c}
\mathbf{A}_i\\
\dot{\mathbf{A}}_i
\end{array}\right]
\mathbf{u}_i +\left[\begin{array}{c}
\textbf{0} \\
\mathbf{A}_i
\end{array}\right] \dot{\mathbf{u}}_i
\end{equation}
 where $\mathbf{p}_i=[x_i,y_i]^\text{T}$ and $\mathbf{\dot p}_i=\mathbf{v}_i=[v_i,w_i]^\text{T}$ are the position and the velocity of the control point (usually the mass center) in the robot $i$, respectively.
 $\mathbf{\dot v}_i=\mathbf{a}_i$ denotes the vehicle acceleration.
 $\mathbf{A}_i$ is the Jacobian matrix, and $\dot{\mathbf{A}}_i$ is the time derivative of $\mathbf{A}_i$. $\mathbf{u}_i$ is the wheel velocity. In this paper, $\dot{\mathbf{u}}_i$, \emph{i.e.}, the wheel acceleration, is chosen as the control input. We will construct the safety barrier function in acceleration level, otherwise, the WMRs could only just their velocities to zero instantaneously in emergency braking situations to avoid collisions. $\dot{\mathbf{u}}_i$ is limited by $\dot{\mathbf{u}}_i^+$ and $\dot{\mathbf{u}}_i^-$, $\dot{\mathbf{u}}_i^-\leq\dot{\mathbf{u}}_i\leq\dot{\mathbf{u}}_i^+$.
\vspace{-15pt}
\subsection{Problem Formulation}
Considering $N$ WMRs described by Eq. (\ref{eqn.kinematic}) enclosed
in a shared workspace, our objective is to design the control law
to compute $\dot{\mathbf{u}}_i$ for every robot so that
minimizing the difference between the actual position $\mathbf{p}_i$ and the desired position $\mathbf{p}_{d_i}$ while satisfying both $\dot{\mathbf{u}}_i\in [\dot{\mathbf{u}}_i^-,\dot{\mathbf{u}}_i^+]$ and
the CA constraint.
\vspace{-15pt}
\subsection{Control Barrier Function}
The basic idea of CBF-based safety control method is to define a safe set $\mathcal{C}$, \emph{i.e.}, the system states $\mathbf{x}$ having no collisions, and then use the class-$\mathscr{K}$ function to formally guarantee the forward invariance of the safe set, \emph{i.e.}, if $\mathbf{x}(0)\in \mathcal{C}$ , then $\mathbf{x}(t)\in \mathcal{C}$ for all $t\geq0$.

Specifically, the safe set is defined as the superlevel set of a continuously differentiable function $h$:
\begin{equation}
\mathcal{C}=\left\{\mathbf{x} \in \mathbb{R} \mid h(\mathbf{x}) \geq 0\right\}
\end{equation}
we say $h$ is a CBF \cite{ACC2021} if $\partial h/\partial \mathbf{x} \neq0$ for all $\mathbf{x}\in\partial \mathcal{C}$ and there exists an extended class-$\mathscr{K}$ function $\psi(\cdot)$ such that $h$ satisfies
\begin{equation}\label{eqn.bf}
\dot{h}(\mathbf{x}) \geq-\psi(h(\mathbf{x}))
\end{equation}
The extended class--$\mathscr{K}$ function $\psi(\cdot)$ regulates the rate of the system states converge to the boundary of $\mathcal{C}$. \emph{E.g}, $\psi(h(\mathbf{x})$ is chosen as $\kappa h^3$ \cite{Safety2017TRO}, $h^{-1}$ \cite{IFAC2015}. Different choices of $\psi(\cdot)$ lead to different behaviors near the boundary. In this paper, the task is chosen as the TT which requires the position of the robot to converge to a desired constant position, therefore the ZCBF is adopted, \emph{i.e.}, $\psi(h(\mathbf{x}))=\kappa h$, $\kappa >0$.
\vspace{-20pt}
\subsection{CA Function}
There are two collision risk assessment functions. One is based on the Euclidean distance, the corresponding safe set $\mathcal{C}_{ij}$ is set as (taking the pairwise WMRs $i$, $j$ as an example)
\begin{equation}\label{eqn.n1}
h_{ij}\geq 0, h_{ij}=||\mathbf{p}_{ij}||- d_{\text{safe}}
\end{equation}
where $\mathbf{p}_{ij}=\mathbf{p}_{i}-\mathbf{p}_{j}$ denotes the relative distance of the pairwise WMRs $i,j$, $d_{\text{safe}}$ is the safety threshold defined by the user, and $||\cdot||$ is the Euclidean norm.
The other is that the maximum relative braking force is considered in the safety constraint $\mathcal{C}_{ij}$, \emph{i.e},

\begin{equation}\label{eqn.n2}
h_{ij}\geq0, h_{ij}=||\mathbf{p}_{ij}||+d_{\text{brake}}- d_{\text{safe}}
\end{equation}
where $d_{\text{brake}}$ denotes the braking distance,
$d_{\text{brake}}=\frac{\dot{\left\|\mathbf{p}_{ij}\right\|}^2}{2\left(a_i^++a_j^+\right)}$.
$\dot{||\mathbf{p}_{ij}||}=\frac{\mathbf{p}_{ij}^\text{T}\mathbf{v}_{ij}}{||\mathbf{p}_{ij}||}$ is the time derivative of $||\mathbf{p}_{ij}||$, denotes the normal component of the relative velocity $\mathbf{v}_{ij}$, $\mathbf{v}_{ij}=\mathbf{v}_{i}-\mathbf{v}_{j}$.
Superscript $\cdot^\text{T}$ denotes the transpose of the matrix, $a_i^+,a_j^+$ are the maximum acceleration of WMRs $i$ and $j$, respectively. Eq. (\ref{eqn.n2}) can be rewritten as
\begin{equation}\label{eqn.n4}
h_{ij}\geq 0,~~
h_{ij}=\begin{aligned}
\sqrt{2\left(a_i^++a_j^+\right)\left(\left\| \mathbf{p}_{i j}\right\|-d_{\text{safe}}\right)}+\dot{\left\|\mathbf{p}_{ij}\right\|}
\end{aligned}
\end{equation}

In this paper, we restrict our analysis to circular WMRs with the sensing range being $D$. We model all WMRs by the smallest enclosing disk of radius $r_i$, $i=1, 2, \cdots, N$. For Eq. (\ref{eqn.n1}), the CA strategy
\begin{equation}\label{eqn.nn5}
  -2\mathbf{p}_{ij}^\text{T}\mathbf{v}_{ij}\leq \varsigma(||\mathbf{p}_{i j}||^2-d_{\text{safe}}^2)
  \end{equation}
is conservative. As pointed in \cite{Li2023DV-SOATT,rodriguez2019cooperative}, the CA module does not have to be activated when the pairwise WMRs move away from each other. To this, a binary parameter $\beta_{ij}$ that characters the trend of assembling or departing is introduced in Eq. (\ref{eqn.nn5}) \cite{Li2023DV-SOATT}, the pairwise collision-free constraint is formulated as
  \begin{equation}\label{eqn.n5}
  -2\beta_{ij}\mathbf{p}_{ij}^\text{T}\mathbf{v}_{ij}\leq \beta_{ij}\varsigma(||\mathbf{p}_{i j}||^2-d_{\text{safe}}^2)
  \end{equation}
  where $\varsigma>0$ is a control parameter,
\begin{equation}\label{eqn.n6}
\left\{
\begin{aligned}
&\beta_{ij}=1~~~~\text{if}~~\mathbf{p}_{ij}^\text{T}\mathbf{v}_{ij}<0~~\text{and}~~||\mathbf{p}_{ij}||<d_{\text{safe}}\\
&\beta_{ij}=0~~~~\text{others}
\end{aligned}
\right.
\end{equation}
In \cite{rodriguez2019cooperative}, the CA's conservatism is reduced by modifying $d_\text{safe}$ from a fixed value as
\begin{equation}\label{eqn.n7}
d_\text{safe}=\frac{D+r}{2}+\frac{D-r}{\pi}\tan^{-1}(-\varrho(\mathbf{p}_{ij}^\text{T}\mathbf{v}_{ij})+\sigma)
\end{equation}
where $r=r_i+r_j$. $\varrho>0$ and $\sigma\in\mathbb{R}$ are two constant design parameters.
We would give the comparison on CA's invasiveness between Eq. (\ref{eqn.n4}), Eq. (\ref{eqn.nn5}), Eq. (\ref{eqn.n5}), Eq. (\ref{eqn.n7}) and our method in Section. \ref{sec.5}.
\section{AA-SOATT Method}\label{sec.3}
In this paper, we give our CA strategy, TT strategy and the deadlock avoidance strategy.
\subsubsection{Collision Avoidance}
 By a double integrator, the pairwise safety set $\mathcal{C}_{ij}$ assisted with the extended class-$\mathscr{K}$ function could be defined as:
\begin{equation}\label{eqn.7}
h_{ij}\geq 0,~~
h_{ij}=\dot{\left\|\mathbf{p}_{ij}\right\|}+\kappa_{ij_1}\left(\left\|\mathbf{p}_{i j}\right\|-d_{\text {safe }}\right)
\end{equation}
$\kappa_{ij_1}>0$ is used to regulate the rate of $\left\|\mathbf{p}_{i j}\right\|\geq d_{\text {safe}}$.

To guarantee the forward invariance of the pairwise safety set Eq. (\ref{eqn.7}), we construct another ZCBF
\begin{equation}\label{eqn.7a}
\dot h_{ij}+\kappa_{ij_2}h_{ij}\geq 0
\end{equation}
The extended class-$\mathscr{K}$ function $\kappa_{ij_2}h_{ij}$ is used to regulate the rate of $h_{ij}\geq 0$, with $\kappa_{ij_2}>0$. The time derivative of $h_{ij}$ is
\begin{equation}\label{eqn.7n}
\begin{aligned}
\dot h_{ij}=\frac{(||\mathbf{v}_{ij}||^2+\mathbf{p}_{ij}^\text{T}\mathbf{\dot v}_{ij})||\mathbf{p}_{ij}||-\mathbf{p}_{ij}^\text{T}\mathbf{v}_{ij}\dot {||\mathbf{p}_{ij}||}}{||\mathbf{p}_{ij}||^2}
+\kappa_{ij_1}\dot{||\mathbf{ p}_{ij}||}
\end{aligned}
\end{equation}
Combined Eq. (\ref{eqn.7a}), Eq. (\ref{eqn.7n}) with Eq. (\ref{eqn.kinematic}), we obtain
\begin{equation}\label{eqn.7c}
\begin{aligned}
&\frac{||\mathbf{v}_{ij}||^2+\mathbf{p}_{ij}^\text{T}(\mathbf{A}_i\mathbf{\dot u}_i+\mathbf{\dot A}_i\mathbf{u}_i-
\mathbf{A}_j\mathbf{\dot u}_j+\mathbf{\dot A}_j\mathbf{u}_j)}
{||\mathbf{p}_{ij}||}
-\frac{(\mathbf{p}_{ij}^\text{T}\mathbf{v}_{ij})^2}{||\mathbf{p}_{ij}||^3}\\
&+(\kappa_{ij_1}+\kappa_{ij_2})\frac{\mathbf{p}_{ij}^\text{T}\mathbf{v}_{ij}}{||\mathbf{p}_{ij}||}
+\kappa_{ij_1}\kappa_{ij_2}(||\mathbf{p}_{ij}||-d_{\text{safe}})\geq 0
\end{aligned}
\end{equation}
Based on Eq. (\ref{eqn.7c}), further, the following collision-free motion constraint between the neighboring pairwise WMRs $i,j$ is obtained:
\begin{equation}\label{eqn.8}
\begin{aligned}
- \mathbf{p}_{i j}^\text{T} \textbf{C}\mathbf{A}_{\text{blk}} \mathbf {\dot u}&  \leq -\frac{\left( \mathbf{p}_{i j}^\text{T}  \mathbf{v}_{i j}\right)^2}{\left\| \mathbf{p}_{i j}\right\|^2}+\left\| \mathbf{v}_{i j}\right\|^2 \\
& +(\kappa_{ij_1}+\kappa_{ij_2})\mathbf{p}_{i j}^\text{T} \mathbf{v}_{i j}\\
& +\kappa_{ij_1}\kappa_{ij_2}||\mathbf{p}_{i j}||(||\mathbf{p}_{i j}||-d_{\text{safe}})\\
&+ \mathbf{p}_{i j}^\text{T}C\mathbf{\dot A}_{\text{blk}}\mathbf{u}
, ~\forall i,j\in N, ~i \neq j, ~j\in \mathscr{N}_i
\end{aligned}
\end{equation}
where $\textbf{C}=[\textbf{0},\cdots,\overbrace{\textbf{I}}^{\text{Robot} i},\cdots,\overbrace{-\textbf{I}}^{\text{Robot}j},\cdots,\textbf{0}]$, $\mathbf{A}_{\text{blk}}$ ($ \mathbf{\dot A}_{\text{blk}})$ is a diagonal matrix
which diagonally places $[\mathbf{A}_1,\cdots,\mathbf{A}_i,\cdots,$
$\mathbf{A}_j,\cdots,\mathbf{A}_N]$$([\mathbf{\dot A}_1,\cdots,\mathbf{\dot A}_i,\cdots,\mathbf{\dot A}_j,\cdots,\mathbf{\dot A}_N])$, $\mathbf{u}=[\mathbf{u}_1,\cdots$
$,\mathbf{u}_i,\cdots,\mathbf{u}_j,\cdots,\mathbf{u}_N]^\text{T}.$ $\mathscr{N}_i$ denotes the neighbour set of the robot $i$,
 \emph{i.e}, when $||\mathbf{p}_{ij}||\leq D$, $j\in \mathscr{N}_i$.

{\it{Theorem 1}}\cite{ccdc2014,2015Robustness,IFAC2015}: Given the pairwise safety set defined by Eq. (\ref{eqn.7}), the ZCBF function $h_{ij}$, and the initial state $h_{ij}(0)=(\mathbf{p}(0), \mathbf{v}(0))\in h_{ij}$, the pairwise safety set $h_{ij}$ is forward invariant, \emph{i.e}, solution that starts in Eq. (\ref{eqn.7}) stays in the safety set for all time.

\text{Proof:}
 Construct $S_{ij}=h_{ij}(0)e^{-\kappa_{ij_2}t}$ with $S_{ij}(0)=h_{ij}(0)$, where $h_{ij}(0)$ denotes the initial value of $h_{ij}$. The time derivative of $S_{ij}$ is $\dot S_{ij}=-\kappa_{ij_2}h_{ij}(0)e^{-\kappa_{ij_2}t}=-\kappa_{ij_2}S_{ij}$.
 Based on the comparison principle, we have $h_{ij}\geq S_{ij}$ for all $t\geq 0$. Therefore,
 \begin{equation}
 h_{ij}\geq h_{ij}(0)e^{-\kappa_{ij_2}t}~~~~~~~~~~~~~~
\end{equation}
Considering $h_{ij}(0)\geq 0$, and $h_{ij}(0)=0$ only if $h_{ij}(0)=0$, therefore
 \begin{equation}
 h_{ij}\geq 0~~\text{for}~~\text{all}~~t\geq 0~~~~~~~~
\end{equation}
The forward invariance of the safety set $h_{ij}$ is guaranteed.
Therefore, the pairwise collision-free constraint is satisfied.
$\hfill\blacksquare$

{\it Remark 1}: Extend to the MWMRs system, the safety set can be defined as the intersection of all possible pairwise safety set. Therefore,
if all possible pairwise safety constraints is satisfied and all WMRs' initial states are within the safety set, based on the forward invariant the pairwise safety set, the whole system is guaranteed to be safe.

Letting $B_{ij}=-\mathbf{p}_{ij}^\text{T} C\mathbf{A}_{\text{blk}}$, $B_{\text{right}_{ij}}=-\frac{\left( \mathbf{p}_{i j}^\text{T}  \mathbf{v}_{i j}\right)^2}{\left\| \mathbf{p}_{i j}\right\|^2}+\left\| \mathbf{v}_{i j}\right\|^2$
$+(\kappa_{ij_1}+\kappa_{ij_2})\mathbf{p}_{i j}^\text{T} \mathbf{v}_{i j}
+\kappa_{ij_1}\kappa_{ij_2}||\mathbf{p}_{i j}||(||\mathbf{p}_{i j}||-d_{\text{safe}})+\mathbf{p}_{i j}^\text{T}C$
$\mathbf{\dot A}_{\text{blk}}\mathbf{u}$, Eq. (\ref{eqn.8}) is rewritten as
\begin{equation}
B_{ij}\mathbf{\dot u}\leq B_{\text{right}_{ij}}
\end{equation}

\subsubsection{Trajectory Tracking}
Similarly, TT strategy defined in the acceleration level can be described as:
\begin{equation}\label{eqn.11}
\mathbf{\ddot p}_{i}-\mathbf{\ddot p}_{d_i}+\kappa_{i_3}(\mathbf{\dot p}_{i}-\mathbf{\dot p}_{d_i})=\kappa_{i_4}
(\mathbf{\dot p}_{i}-\mathbf{\dot p}_{d_i}+\kappa_{i_3}\ell_i),
\end{equation}
where $\ell_i=\mathbf{p}_{i}-\mathbf{p}_{d_i}$ denotes the error between the reference position and the actual position of the robot $i$, and $\kappa_{i_3},\kappa_{i_4}$ is used to regulate the tracking accuracy of $\mathbf{p}_{i}\rightarrow\mathbf{p}_{d_i}$. Combined with Eq. (\ref{eqn.kinematic}), we have
\begin{equation}\label{eqn.12}
\begin{aligned}
\mathbf{A}_{i}\mathbf{\dot u}_{i}=\text{dzr}_i
\end{aligned}
\end{equation}
where $\text{dzr}_i=\mathbf{a}_{d_i}-\mathbf{\dot A}_{i}\mathbf{u}_{i}-(\kappa_{i_3}+\kappa_{i_4})(\mathbf{A}_{i}\mathbf{u}_{i}-\mathbf{v}_{d_i})
-\kappa_{i_3}\kappa_{i_4}\ell_i$.
\subsubsection{Deadlock Avoidance}
Deadlock is rather frequent for MWMRs systems, especially in a crowded environment.
To avoid deadlock, \cite{Safety2017TRO} imposes an additional disturbance on $\text{dzr}_i$, Eq. (\ref{eqn.12}) is rewritten as
\begin{equation}\label{eqn.12a}
\begin{aligned}
\mathbf{A}_{i}\mathbf{\dot u}_{i}=(\textbf{I}-\mathscr{\textbf{Q}})\text{\textbf{dzr}}_i
\end{aligned}
\end{equation}
where the matrix $\mathscr{\textbf{Q}}_i$ is defined as
\begin{equation}\label{eqn.14}
\begin{split}
&\mathscr{\textbf{Q}}_i=\begin{bmatrix}
\cos(q_i)&-\sin(q_i)\\
\sin(q_i)&\cos(q_i)
\end{bmatrix}\in \mathbb{R}^{2\times 2}
\end{split}
\end{equation}
$q_i\in[-180^{\circ},180^{\circ}]$.
In \cite{he2024simultaneous}, the deadlock is prevented from modifying the relative distance $\mathbf{p}_{ij}$ as
\begin{equation}\label{eqn.12c}
\mathbf{p}_{ij}=(\textbf{I}-\mathscr{\textbf{Q}})\mathbf{p}_{ij}
\end{equation}

Different from ways of \cite{Safety2017TRO} and \cite{he2024simultaneous}, in this paper, we investigate influence of other two ways on the invasiveness of CA. Deadlock
a situation that WMRs in the current situation form a symmetrical geometric layout.
As pointed in \cite{grover2021deadlock}, the sum of the attraction force from the target and the repulsive force from the obstacle being zero, results in zero velocity for the robot.
Therefore, we imagine that when the deadlock decision condition is satisfied, making the robot deviate from its preferred trajectory to the left or right to disrupt this equilibrium and avoid deadlock. Specifically,
\begin{itemize}
\item[i)] We first investigate influence of disturbing the preferred velocity $\mathbf{v}_{d_i}$ on the invasiveness of CA, \emph{i.e},
\begin{equation}\label{eqn.12b}
\begin{aligned}
\mathbf{v}_{d_i}=(\textbf{I}-\zeta_i\mathscr{\textbf{Q}})\mathbf{v}_{d_i}
\end{aligned}
\end{equation}
\item[ii)] Except for a way of disturbing the preferred velocity, an auxiliary velocity term is introduced into the error function $\mathbf{A}_{i}\mathbf{u}_{i}-\mathbf{v}_{d_i}+\kappa_{i_3}\ell_i$ of the TT. Then,

\begin{equation}\label{eqn.13}
\begin{aligned}
&\mathbf{A}_{i}\mathbf{\dot u}_{i}=\text{dzr}_i\\
&\text{dzr}_i=\text{dzr}_{1_i}+\text{dzr}_{2_i}\\
\end{aligned}
\end{equation}
where $\text{dzr}_{1_i}=\mathbf{a}_{d_i}-\mathbf{\dot A}_{i}\mathbf{u}_{i}-(\kappa_{i_3}+\kappa_{i_4})(\mathbf{A}_{i}\mathbf{u}_{i}-\mathbf{v}_{d_i})
-\kappa_{i_3}\kappa_{i_4}\ell_i$,
$\text{dzr}_{2_i}=-\zeta_i\mathscr{\textbf{Q}}(\mathbf{A}_{i}\mathbf{u}_{i}-\mathbf{v}_{d_i}+\kappa_{i_3}\ell_i)$.
In our method, control parameter $\zeta_i$ is introduced, which is used to regulate the margin of deviating from the preferred trajectory, with

\begin{equation}\label{eqn.15}
\left\{
\begin{aligned}
&\zeta_i>0~~~~\text{if}~~\eta_{\in i}>0~~\\
&\zeta_i=0~~~~\text{others}
\end{aligned}
\right.
\end{equation}
 $\eta_{\in i}$ represents that the CA needs to be activated for the robot $i$, the details is given in Section \ref{sec.4}.


{\it{Theorem 2}}: TT error of the robot $i$ that subjects to Eq. (\ref{eqn.13}) and Eq. (\ref{eqn.kinematic}) globally converges to zero when
$t\rightarrow\infty$, if $\kappa_{i_4}>\zeta_i$. For the MWMRs system, globally convergence of every WMR's TT error is guaranteed by $\kappa_{i_4}>\zeta_i$.

\text{Proof:} Define
\begin{equation}\label{eqn.a}
\begin{aligned}
&e_i=\mathbf{\dot p}_{i}-\mathbf{\dot p}_{d_i}+\kappa_{i_3}(\mathbf{p}_{i}-\mathbf{p}_{d_i})\\
&\dot e_i=-\kappa_{i_4}e_i-\zeta_i\mathscr{\textbf{Q}}e_i
\end{aligned}
\end{equation}
Construct a Lyapunov function $V_{i_1}=e_i^\text{T}e_i/2$. Obviously, $V_{i_1}\geq0$, and $V_{i_1}=0$ only when $e_i=0$.
Taking the time derivation of $V_{i_1}$ and substituting by (\ref{eqn.a}) yields
\begin{equation}
\dot V_{i_1}=-(\kappa_{i_4}+\zeta_i\mathscr{\textbf{Q}})e_i^2
\end{equation}
To ensure convergence, we need to ensure $\kappa_{i_4}+\zeta_i\mathscr{Q}>0$, \emph{i.e}, $\kappa_{i_4}>-\zeta_i\mathscr{Q}$. Due to $\zeta_i\geq0$, we have
$\mathbf{\dot p}_{i}-\mathbf{\dot p}_{d_i}$ converges to zero when $t\rightarrow \infty$, if $\kappa_{i_4}>\zeta_i$. Define
\begin{equation}\label{eqn.a2}
\begin{aligned}
&\ell_i=\mathbf{p}_{i}-\mathbf{p}_{d_i}\\
&\dot \ell_i=-\kappa_{i_3}\ell_i
\end{aligned}
\end{equation}
Construct a Lyapunov function $V_{i_2}=\ell_i^\text{T}\ell_i/2$. Obviously, $V_{i_2}\geq0$, and $V_{i_2}=0$ only when $\ell_i=0$.
Taking the time derivation of $V_{i_2}$ and substituting by (\ref{eqn.a2}) yields
\begin{equation}
\dot V_{i_2}=-\kappa_{i_3}\ell_i^2
\end{equation}
Because $\kappa_{i_3}>0$, $\dot V_{i_2}\leq0$, and $\dot V_{i_2}=0$ only when $\ell_i=0$. Using LaSalle’s invariant principle \cite{TNNLs2017}, we can conclude that the tracking error $\ell_i=\mathbf{p}_{i}-\mathbf{p}_{d_i}$ globally converges to zero when $t\rightarrow \infty$.

{\it Remark 2}: In this paper, $\mathscr{\textbf{Q}}$, together with $\zeta_i$, is the same for all WMRs. Therefore, for the MWMRs system, $\mathbf{ p}-\mathbf{ p}_d$ converges to zero when $t\rightarrow \infty$, provided that $\kappa_{i_4}>\zeta_i$,
$\mathbf{p}=[\mathbf{p}_1,\cdots,\mathbf{p}_N]^\text{T}$,
$\mathbf{p}_d=[\mathbf{p}_{d_1},\cdots,\mathbf{p}_{d_N}]^\text{T}$.
$\hfill\blacksquare$
\end{itemize}
\vspace{-20pt}
\subsection{AA-SOATT Method}\label{sec.2.3}
\vspace{-5pt}
Based on the CA and TT strategies proposed above, the AA-SOATT method of the robot $i$ is described as a QP minimization formulation:
\begin{subequations}\label{eqn.16}
\begin{align}
&\mathop{\text{min}}\limits_{\mathbf{\dot u_i}}~~~(\mathbf{A}_i\mathbf{\dot u}_i-\text{dzr}_i)^\text{T}(\mathbf{A}_i\mathbf{\dot u}_i-\text{dzr}_i)\label{equ.16a}\\
&~\text{s.t.}~~~~~~~~~\mathbf{\dot u_i}^{-}\le \mathbf{\dot u_i}\le \mathbf{\dot u_i}^{+}~~~~~~~~\label{equ.16c}\\
&~~~~~~~~~~~B_{ij}\mathbf{\dot u}\leq B_{\text{right}_{ij}}
, ~\forall i,j\in N, ~i \neq j, ~j\in \mathscr{N}_i\label{equ.16d}
\end{align}
\end{subequations}
Letting $\text{dzr}=[\text{dzr}_1,\cdots,\text{dzr}_i,\cdots,\text{dzr}_N]^\text{T}$,
$B=[B_{12},\cdots,$
$B_{1N},B_{23},\cdots,B_{(N-1)N}]^\text{T}, B_{\text{right}}=[B_{\text{right}_{12}},\cdots,B_{\text{right}_{1N}},$
$B_{\text{right}_{23}},\cdots,B_{\text{right}_{(N-1)N}}]^\text{T}$, the combined-formed AA-SOATT considering $N$ WMRs is given :
\begin{subequations}\label{eqn.finalAA-SOATT}
\begin{align}
&\mathop{\text{min}}\limits_{\mathbf{\dot u}}~~~(\mathbf{A}_{\text{blk}}\mathbf{\dot u}-\text{dzr})^\text{T}(\mathbf{A}_{\text{blk}}\mathbf{\dot u}-\text{dzr})\label{equ.19a}\\
&~~\text{s.t.}~~~~\mathbf{\dot u}^{-}\leq \mathbf{\dot u}\leq \mathbf{\dot u}^{+}~~~~~~~~~~~~~~~~~~\label{equ.19b}\\
&~~~~~~~~~~~B\mathbf{\dot u}\leq B_{\text{right}}
\end{align}
\end{subequations}
\section{Control law and stability Analyses}\label{sec.4}
\vspace{-10pt}
In this paper, the control law is built on the Lagrange multipliers method.
Define the following Lagrange function of Eq. (\ref{eqn.finalAA-SOATT}):
\begin{equation}\label{eqn.20}
\begin{split}
L(\mathbf{\dot u},\eta)&=(\mathbf{A_{\text{blk}}}\mathbf{\dot u}-\text{dzr})^\text{T}(\mathbf{A_{\text{blk}}}\mathbf{\dot u}-\text{dzr})\\
&+\eta^\text{T}(B\mathbf{\dot u}-B_{\text{right}})
\end{split}
\end{equation}
where $\eta\in \mathbb{R}^{(N^2-N)/2}$ is Lagrange multiplier.
Optimal solution of Eq. (\ref{eqn.finalAA-SOATT}) corresponds to the saddle point of Lagrangian function Eq. (\ref{eqn.20}).
 To settle down at the saddle point of Eq. (\ref{eqn.20}), based on the projection theorem and the Karush-Kuhn-
Tucker condition\cite{TNNLS2020},
we have
\begin{subequations}\label{eqn.21}
\begin{align}
&\varepsilon\frac{\partial \mathbf{\dot u}}{\partial t}=-(\mathbf{\dot u}-\mathbf{\tilde{\dot u}})\\
&\varepsilon\frac{\partial \eta}{\partial t}=-(\eta-\tilde{\eta})
\end{align}
\end{subequations}
where $\varepsilon>0$ is a time scaling factor, which is used to regulate the convergence rate of $\mathbf{\dot u}\rightarrow\mathbf{\tilde{\dot u}}$ and $\eta\rightarrow\tilde{\eta}$.
$\mathbf{\tilde{\dot u}}=\varphi(\mathbf{\dot u}-\partial L/\partial \mathbf{\dot u})$, $\tilde{\eta}=\max(0,\psi(B\mathbf{\dot u}-B_{\text{right}}+\eta))$, therefore,  Eq. (\ref{eqn.21}) is recast as
\begin{subequations}\label{eqn.22}
\begin{align}
&\varepsilon\mathbf{\ddot u}=-\mathbf{\dot u}+\varphi(\mathbf{\dot u}-2\mathbf{A}_{\text{blk}}^\text{T}(\mathbf{A_{\text{blk}}}\mathbf{\dot u}-\text{dzr})-B^\text{T}\eta)\\
&\varepsilon\dot\eta=\max(0,\psi(B\mathbf{\dot u}-B_{\text{right}}+\eta))-\eta~~~~\label{eqn.22b}
\end{align}
\end{subequations}
Eq. (\ref{eqn.22b}) ensures that $\eta\geq0$. When $\eta_{\in i}>0$ associated with the robot $i$, the CA module of the robot $i$ would be activated. Algorithm 1 gives the determination of $\eta_{\in i}$ .
Bound constraint Eq. (\ref{equ.19b}) is mapped into the piecewise linear projection function $\varphi(\cdot)$.

{\it Theorem 3:} The designed control law Eq. (\ref{eqn.22}) is globally convergent to a unique equilibrium point and is globally asymptotically stable in the Lyapunov sense.

Proof: Eq.(\ref{eqn.22}) can further be rewritten as
\begin{align}\label{RNN2}
\begin{small}
\varepsilon\left[\begin{matrix}\mathbf{\ddot u}\\ \dot\eta\end{matrix}\right]=\left[\begin{matrix}
-\mathbf{\dot u}+\varphi(\mathbf{\dot u}-2\mathbf{A}_{\text{blk}}^\text{T}(\mathbf{A}_{\text{blk}}\mathbf{\dot u}-\text{dzr})-
B^\text{T}\eta),\\
\max(0, \psi(B\mathbf{\dot u}-B_{\text{right}}+\eta))-\eta,~~\\
\end{matrix}
\right]
\end{small}
\end{align}
Let $\chi=[\mathbf{\dot u},\eta]^\text{T}$. Eq. (\ref{RNN2}) is recast as
\begin{align}\label{RNN3}
\varepsilon \dot \chi=-\chi+\phi_{\Lambda}(\chi-\mathcal{H}(\chi))
\end{align}
where
\begin{align}\label{RNN4}
\mathcal{H}(\chi)=\left[\begin{matrix}
2\mathbf{A}_{\text{blk}}^\text{T}(\mathbf{A}_{\text{blk}}\mathbf{\dot u}-\text{dzr})+B^\text{T}\eta),\\
-B\mathbf{\dot u}+B_{\text{right}},
\end{matrix}
\right]
\end{align}
$\Lambda=\{\chi\in\mathbb{R}^{Nm+(N^2-N)/2}|\chi^-\leq \chi\leq \chi^+\}$, the projection operator $\phi_{\Lambda}=[\varphi,\psi]^\text{T}$, and

\begin{equation*}
\begin{split}
&\chi^-=\begin{bmatrix}
 \mathbf{\dot u}^-\in \mathbb{R}^{Nm}~~~~~~\vspace{1ex} \\
 0\in \mathbb{R}^{(N^2-N)/2}\\
\end{bmatrix}
\end{split},~~
\begin{split}
&\chi+=\begin{bmatrix}
  \mathbf{\dot u}^+\in \mathbb{R}^{Nm}~~~~~~~~\vspace{1ex} \\
 +\infty\in \mathbb{R}^{(N^2-N)/2}
\end{bmatrix}
\end{split}
\end{equation*}
Gradient of $\mathcal{H}(\chi)$ is
\begin{equation}\label{RNN4}
\nabla \mathcal{H}(\chi)=\left[\begin{matrix}
2\mathbf{A}_{\text{blk}}^\text{T}\mathbf{A}_{\text{blk}}&B^\text{T}\\
-B&0
\end{matrix}\right]
\end{equation}
Based on \cite{TAC2004}, $\nabla\mathcal{H}(\chi)$ is positive definite.
The equilibrium condition of Eq. (\ref{RNN3}) is
\begin{equation}\label{RNN5}
\phi_{\Lambda}(\chi-\mathcal{H}(\chi))=\chi.
\end{equation}
Eq. (\ref{RNN5}) is equivalent to
\begin{equation}
\mathcal{H}(\chi^*)^\text{T}(\chi-\chi^*)\geq 0,
\end{equation}
where $\chi^*$ is the equilibrium point of Eq. (\ref{RNN3}).
Assume that $\chi_1$ and $\chi_2$ are two equilibrium points of (\ref{RNN3}), we have
\begin{equation}
\left\{
\begin{aligned}
\mathcal{H}(\chi_1)^\text{T}(\chi_2-\chi_1)\geq 0,\\
\mathcal{H}(\chi_2)^\text{T}(\chi_1-\chi_2)\geq 0.
\end{aligned}
\right.
\end{equation}
Then it is obtained that
\begin{equation*}
\begin{split}
(\mathcal{H}(\chi_2)-\mathcal{H}(\chi_1))^\text{T}(\chi_1-\chi_2)\geq 0\\
\end{split}
\end{equation*}
Due to
\begin{equation}
\nabla \mathcal{H}+(\nabla \mathcal{H})^\text{T}=\left[\begin{matrix}
4\mathbf{A}_{\text{blk}}^\text{T}\mathbf{A}_{\text{blk}}&0&0\\
0&0&0\\
0&0&0
\end{matrix}
\right],
\end{equation}
$\nabla \mathcal{H}(\bullet)$ is strictly monotone, therefore, $(\chi_1-\chi_2)^\text{T}(\mathcal{H}(\chi_2)-\mathcal{H}(\chi_1))=0$. This implies that $\chi_1=\chi_2$. Consequently, the control law Eq. (\ref{eqn.22}) has a unique equilibrium point. Stability and global convergence proof are similar to our previous work \cite{Li2023DV-SOATT}, thus the proof is omitted.
$\hfill\blacksquare$
\begin{algorithm}\label{A.2}
 \caption{Determination of $\eta_{\in i}$.}
 \begin{algorithmic}[1]
\FOR{$i=1:N-1$}
\FOR{$j=i+1:N$}
 \STATE
$\alpha=(i-1)(N-1)+j-i-(i-2)(i-1)/2$.\\
\IF{$\eta[\alpha]>0$}
  \STATE
  $\zeta_i,~\zeta_j\leftarrow >0$.
  \ELSE
  \STATE
  $\zeta_i,~\zeta_j\leftarrow 0$.
  \ENDIF\\
\ENDFOR\\
\ENDFOR\\
\end{algorithmic}
 \end{algorithm}

\begin{table*}\centering
 \caption{\label{tab1}Comparison of TT strategies.}
  \renewcommand\arraystretch{1.1}
  \setlength{\tabcolsep}{1.4mm}{
\begin{tabular}{|c|c|c|c|c|c|c|c|c|c|}
\hline Trajectory &
\multicolumn{3}{c|}{RMSE}
 &
\multicolumn{3}{c|}{MAE}
 &
\multicolumn{3}{c|}{Standard Deviation}
\\
\hline  & \begin{tabular}{c}
SBC \\$\kappa_{i_3}=10$
\end{tabular}
& \begin{tabular}{c}
SBC \\$\kappa_{i_3}=80$
\end{tabular}
&\begin{tabular}{c}
Ours \\$\kappa_{i_3}=1$
\end{tabular}
 &
\begin{tabular}{c}
SBC \\$\kappa_{i_3}=10$
\end{tabular}
& \begin{tabular}{c}
SBC \\$\kappa_{i_3}=80$
\end{tabular}
&\begin{tabular}{c}
Ours \\$\kappa_{i_3}=1$
\end{tabular}
 &
\begin{tabular}{c}
SBC \\$\kappa_{i_3}=10$
\end{tabular}
& \begin{tabular}{c}
SBC \\$\kappa_{i_3}=80$
\end{tabular}
&\begin{tabular}{c}
Ours \\$\kappa_{i_3}=1$
\end{tabular}
\\
\hline Straight &0.2000 &0.0259&\textbf{0.0267}& 0.1987&0.0254&\textbf{0.0203}&0.0230 &0.0051&\textbf{0.0173}  \\
\hline
Sine &0.4923 &0.0657&\textbf{0.0640}&0.4857 &0.0629&\textbf{0.0372}&0.0799 &0.0191&\textbf{0.0521}\\
\hline
Circle &0.4108 &0.0568&\textbf{0.0619}&0.4056&0.0549 &\textbf{0.0404}&0.0651&0.0144&\textbf{0.0468}\\
\hline
\end{tabular}}
\end{table*}

\begin{figure*}[]
  \centering
\setlength{\abovecaptionskip}{-0.1cm}   
\subfigcapskip=-4pt
\begin{minipage}[]{7.5in}
\vspace{-10pt}
\hspace{-20pt}
\subfigure[]{\includegraphics[scale=0.1176]{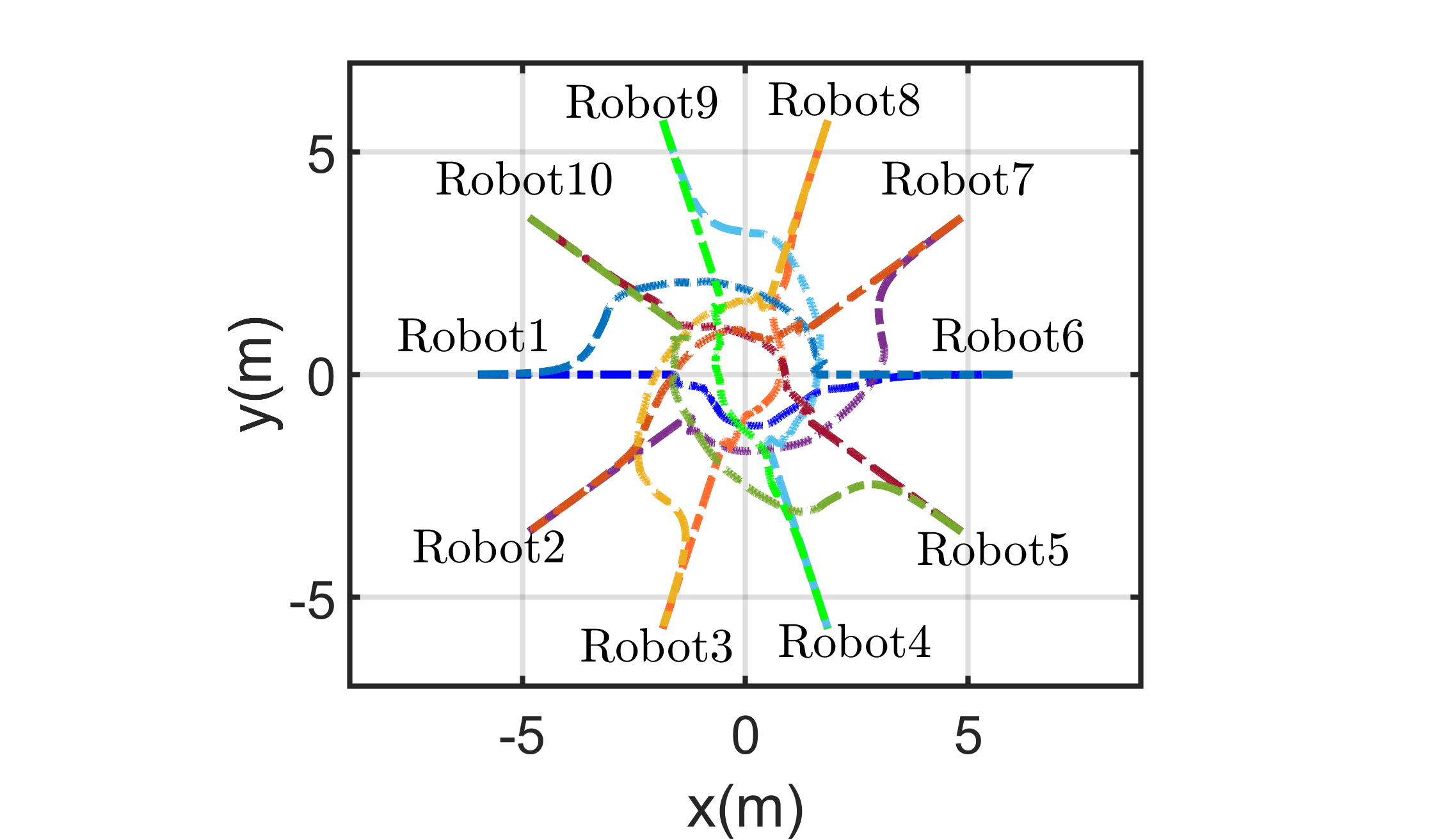}}
\hspace{-28pt}
\subfigure[]{\includegraphics[scale=0.1176]{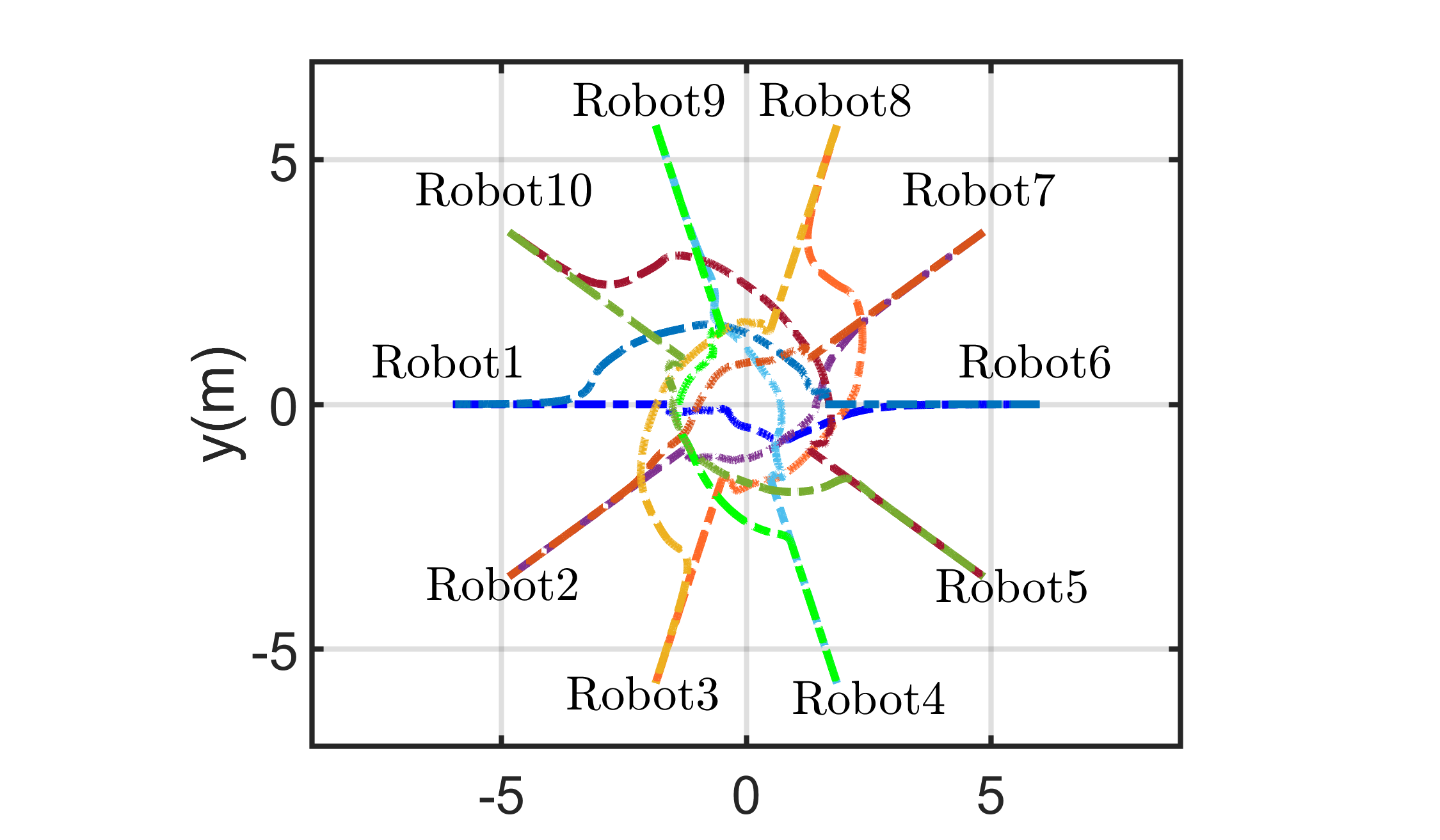}}
\hspace{-28pt}
\subfigure[]{\includegraphics[scale=0.1176]{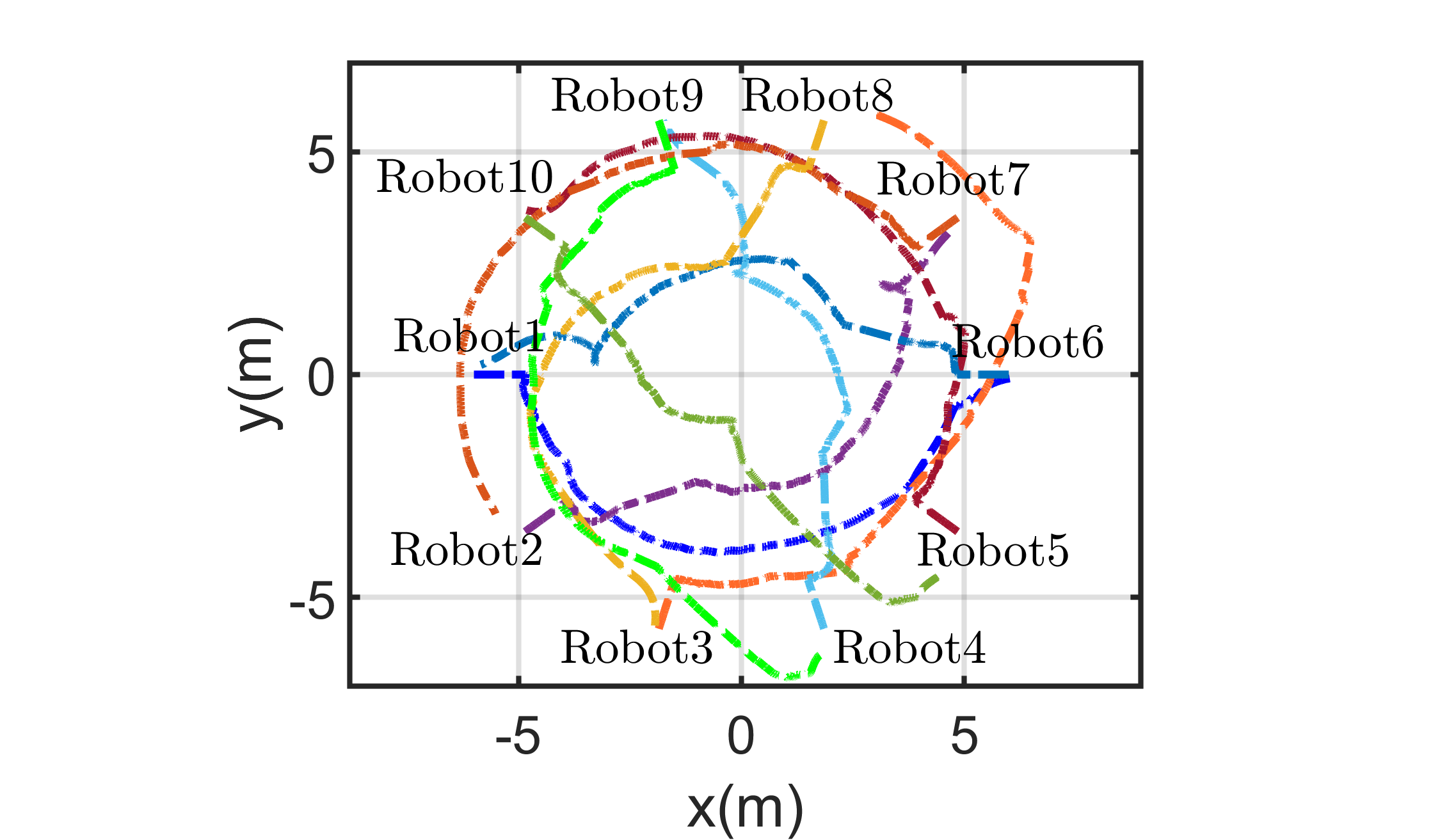}}
\hspace{-28pt}
\subfigure[]{\includegraphics[scale=0.1176]{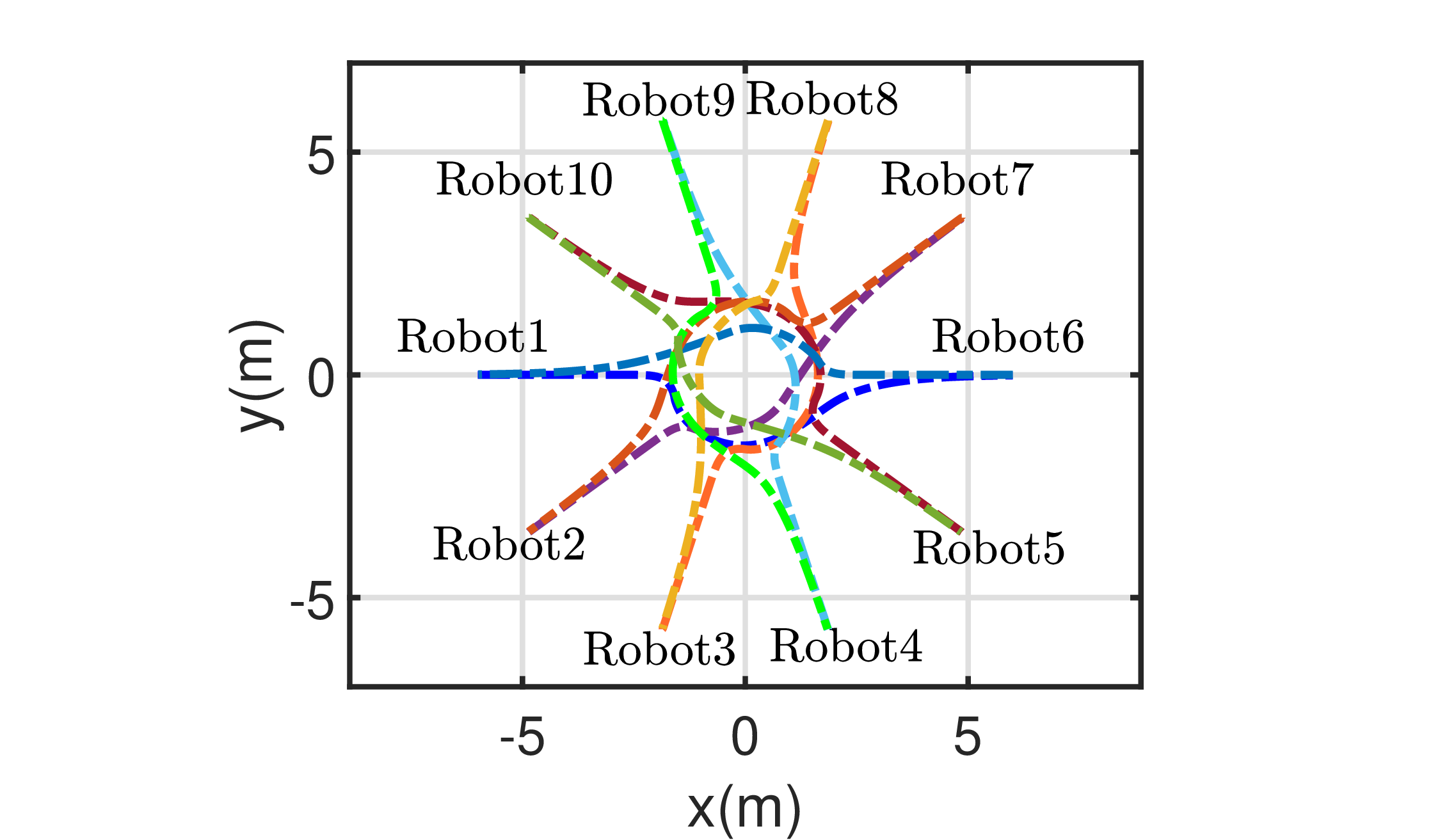}}
\hspace{-28pt}
\subfigure[]{\includegraphics[scale=0.1176]{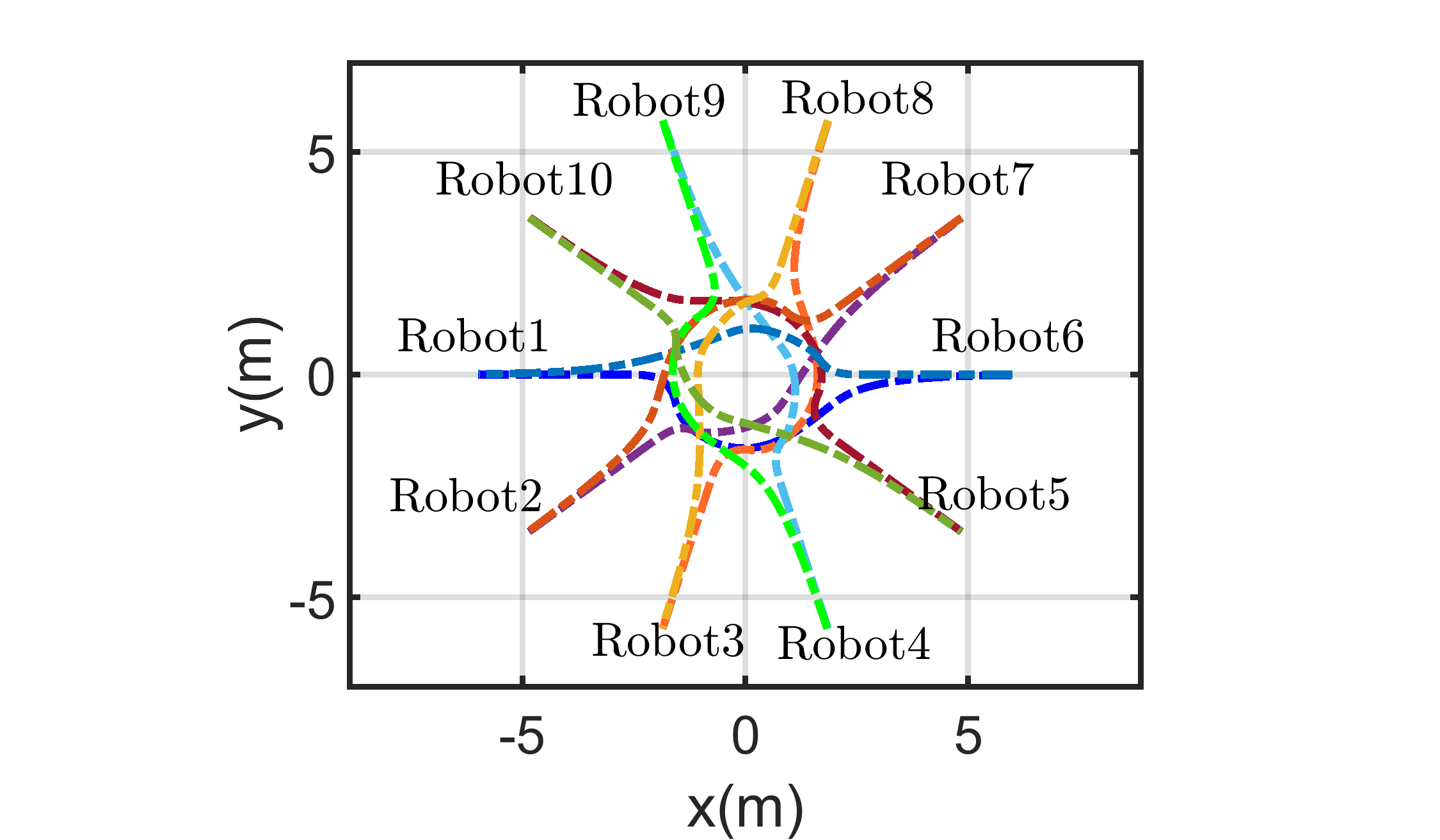}}
\end{minipage}
 \caption{\justifying An illustration of the trajectories generated by Eq. (\ref{eqn.n4}), Eq. (\ref{eqn.n5}), Eq. (\ref{eqn.n7}), Eq. (\ref{eqn.nn5}) and our method, respectively.}\label{fig.reviewacc}
 \end{figure*}
\section{Comparison and Validation}\label{sec.5}
In this part, we would illustrate effectiveness of our method, and superiority compared to the existing methods.
The time step and $\varepsilon$ are set as $0.005$, the total time is $12\mathrm{s}$, unless noted differently.
All $\eta(0), \mathbf{v}(0),\mathbf{u}(0),\mathbf{\dot u}(0)$ are initialized as $0$, and $\theta(0)=\theta_d$. $\theta_d$ denotes the desired heading angle.
The CA's invasiveness is evaluated using the intervention time and proximity of the actual trajectory on the desired trajectory.
Intervention time is the duration from the start to the end of a WMR's CA action. Three metrics (root mean square error (RMSE), mean average error (MAE) and standard deviation) are used to quantify proximity of the actual trajectory on the desired trajectory, and lower values are desirable.
\subsubsection{Comparison of TT strategies}
In the SBC method \cite{Safety2017TRO}, the nominal controller is set as $\hat{u}_i=\kappa_{i_3}(\mathbf{p}_{i}-\mathbf{p}_{d _i})-\kappa_{i_4}\mathbf{v}_i$. When not consider the WMR's kinematics, the nominal controller in our method is $\hat{u}_i=\mathbf{a}_{d_i}-(\kappa_{i_3}+\kappa_{i_4})*(\mathbf{v}_i-\mathbf{v}_{d _i})-\kappa_{i_3}\kappa_{i_4}(\mathbf{p}_{i}-\mathbf{p}_{d_ i})$. Under different $\kappa_{i_3}$ and $\kappa_{i_4}=4$,
Table. \ref{tab1} gives comparison between our TT strategy and one in SBC \cite{Safety2017TRO}, by tracking three common trajectories.
Following Table. \ref{tab1}, we observe that compared to our method, the SBC method obtains level-pegging results depending on a larger control parameter $\kappa_{i_3}$. However, large control parameter may lead to overshoot, with a unstable control behavior. Due to the additional consideration of tracking the preferred velocity and acceleration in our strategy, we achieved the better TT results than the SBC, based on small $\kappa_{i_3}$.

\begin{table*}
\centering
\caption{Comparison of CA strategies.} \label{tab2}
\begin{threeparttable}
\newcommand{\minitab}[2][l]{\begin{tabular}{#1}#2\end{tabular}}
  \setlength{\tabcolsep}{5mm}{
\begin{tabular}{c|c|c|ccccc}
  \hline
    \multicolumn{2}{c|}{\multirow{2}{*}{Metrics}}&\multirow{2}{*}{Methods}&\multicolumn{5}{c}{Numbers of WMRs}\\
    \cline{4-8}
    \multicolumn{2}{c|}{}&& 20 & 25 & 30 & 35& 39  \\
  \hline
\multirow{6}{*}{\makecell{Proximity on\\Desired \\Trajectory}}
&\multirow{2}{*}{RMSE}& Eq. (\ref{eqn.n4})  & 1.1983 & 1.3005 & 1.4055 &- &- \\
      \cline{3-8}
   & & Ours & \textbf{1.1407} & \textbf{1.2988} & \textbf{1.4381} & \textbf{2.0293}&\textbf{2.9324} \\
      \cline{2-8}
  \cline{2-8}
     &\multirow{2}{*}{MAE}
& Eq. (\ref{eqn.n4})  & 0.8465 &0.9709  & 1.1026 & - &- \\
      \cline{3-8}
   & & Ours &\textbf{ 0.8092} & \textbf{0.9730} & \textbf{1.1288 }& \textbf{1.6680} & \textbf{2.5260}\\
      \cline{2-8}
   \cline{2-8}
    &\multirow{2}{*}{\makecell{Standard\\ Deviation}}
   & Eq. (\ref{eqn.n4})  & 0.8482 & 0.8653 & 0.8717 & - &- \\
      \cline{3-8}
   & & Ours & \textbf{0.8040} & \textbf{0.8604} & \textbf{0.8910} & \textbf{1.1558}&\textbf{1.4894} \\
      \cline{2-8}
      \hline
    \multicolumn{2}{c|}{\multirow{2}{*}{Intervention Time}}
    & Eq. (\ref{eqn.n4})  & 7.4650 & 8.5150 & 10.1200 & - &- \\
      \cline{3-8}
    \multicolumn{2}{c|}{}& Ours & \textbf{7.2850} &\textbf{ 8.8250} & \textbf{10.0150} & \textbf{11.3900}&\textbf{11.9900} \\
      \cline{3-8}
 \hline
\end{tabular}}
\begin{tablenotes}
\item[]"-": the safety constraint is violated.
\end{tablenotes}
\end{threeparttable}
\end{table*}

\begin{figure}
  \centering
  \setlength{\abovecaptionskip}{-0.1cm}   
  \begin{minipage}[]{5.5in}
  \subfigure[]{\includegraphics[scale=0.1175]{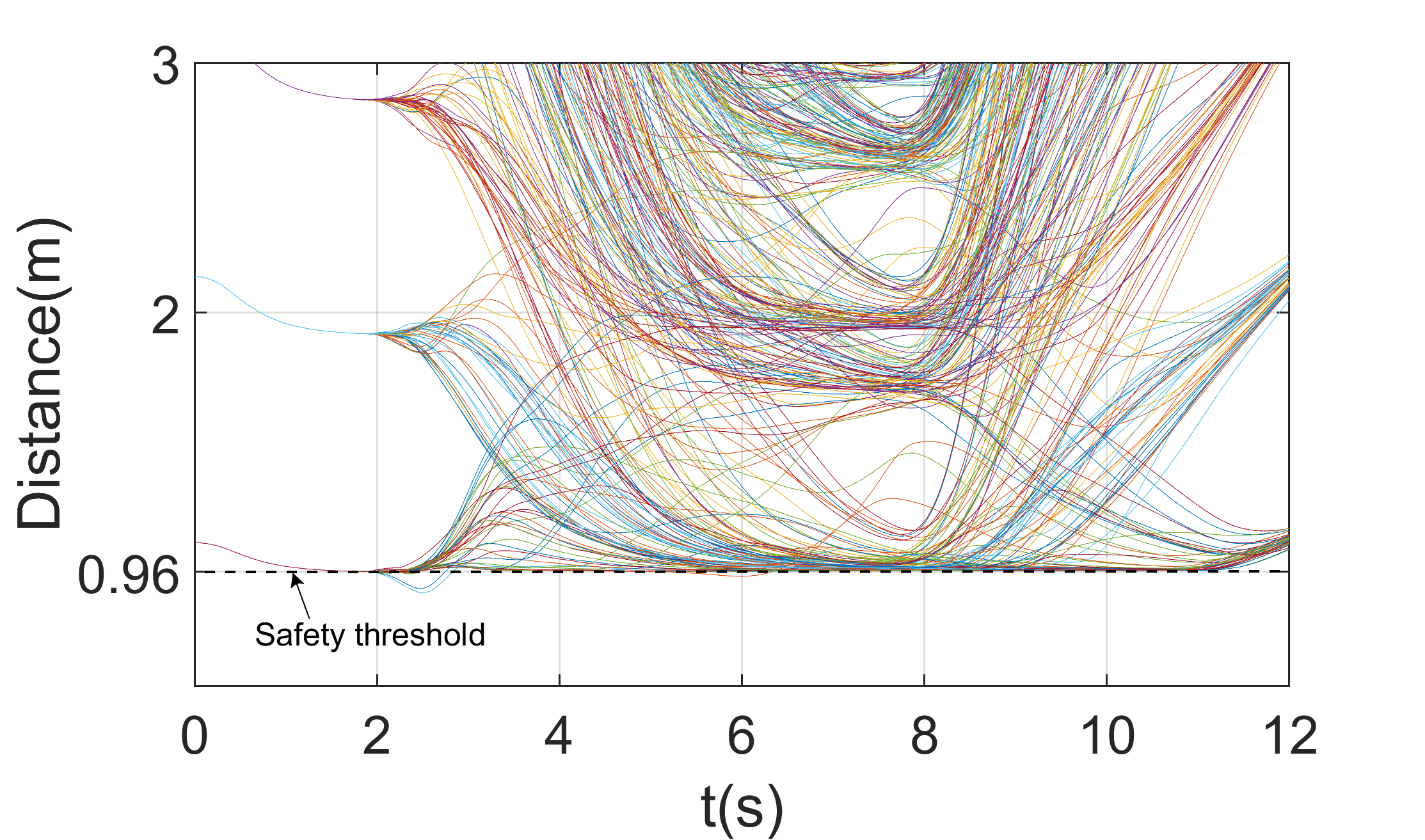}}
  \subfigure[]{\includegraphics[scale=0.1175]{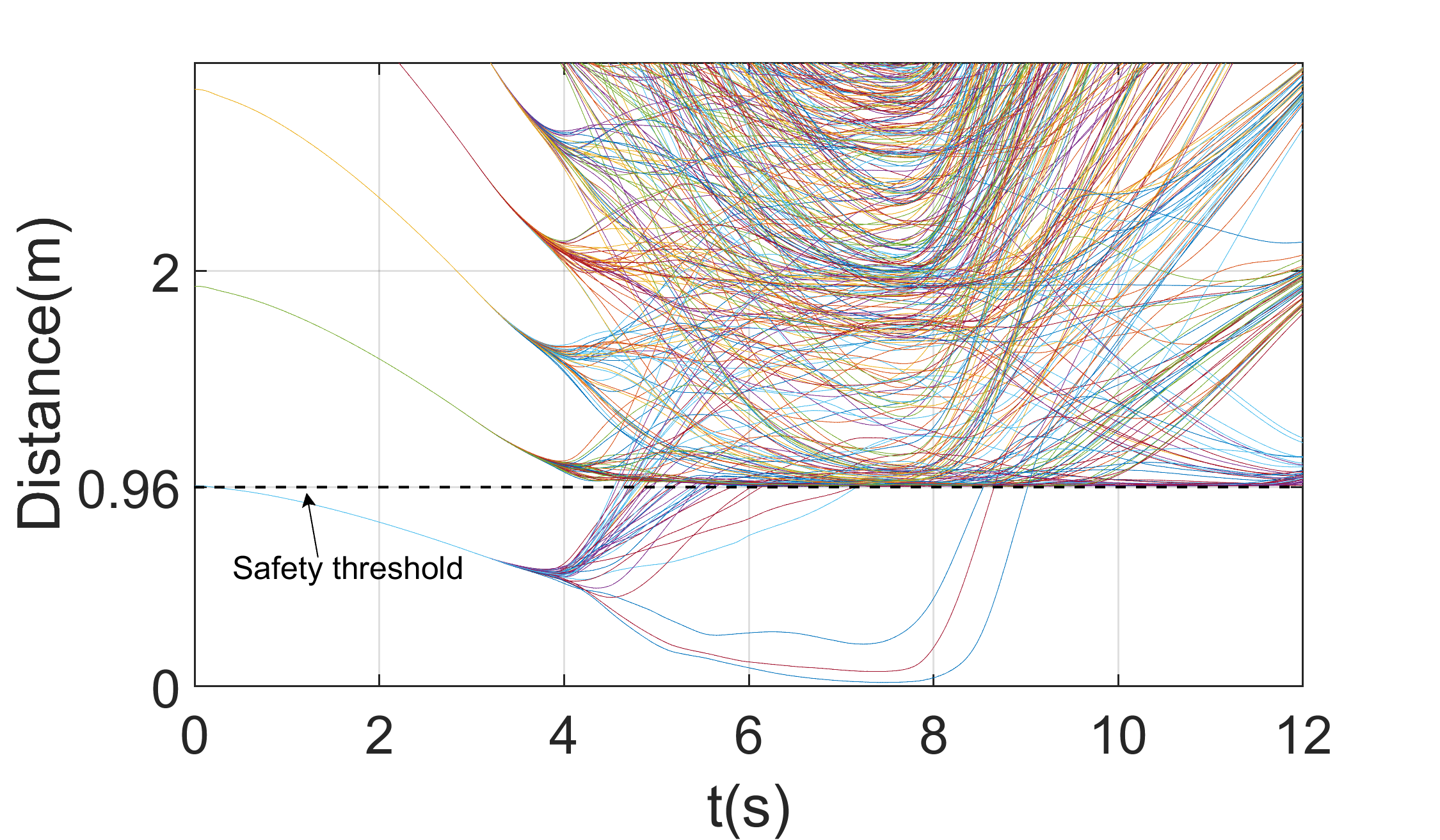}}
  \end{minipage}
  \caption{\justifying Distance profiles generated by Eq. (\ref{eqn.n4})  when the number of WMRs is $35$ and $39$, respectively. The safety threshold is $0.96$. Obviously, the safety constraint is violated in these two examples.}\label{fig2}
\end{figure}
\subsubsection{Comparison of CA strategies}
Considering a circle scenario with the radius being $6$, where $10$ WMRs try to move through the center of a circle to antipodal positions. Fig. \ref{fig.reviewacc} gives an illustration of the trajectories generated by Eq. (\ref{eqn.n4}) \cite{Safety2017TRO}, Eq. (\ref{eqn.n5}) \cite{Li2023DV-SOATT}, Eq. (\ref{eqn.n7}) \cite{rodriguez2019cooperative}, Eq. (\ref{eqn.nn5}) and method proposed by this paper, respectively. Obviously, Eq. (\ref{eqn.n7}) is overly conservative, as shown in Fig. \ref{fig.reviewacc}(c), ten WMRs deviate from their preferred trajectory prematurely. The performance index (Intervention Time, RMSE, MAE, Standard Deviation) is $(10.4150,2.7623,2.2492,1.6036)$. One of reasons for this phenomenon is $d_{\text{safe}}$ in Eq. (\ref{eqn.n7}) is always larger than $r_i+r_j$. The performance indexes of Eq. (\ref{eqn.nn5}) and Eq. (\ref{eqn.n5}) are $(4.5500,0.8257,0.4683,0.6801)$ and $(4.5300,0.7062,0.3820,0.5940)$, repesctively. The performance indexes of Eq. (\ref{eqn.n4}) and our method are $(4.3500,0.7689,$
$0.4810,0.5998)$ and $(4.4850,0.7888,0.4982,0.6115)$, resp
ectively.
From perspective of the generated trajectory, Eq. (\ref{eqn.nn5}) and Eq. (\ref{eqn.n5}) are inferior to Eq. (\ref{eqn.nn5}) and our method. Compared to Eq. (\ref{eqn.nn5}), the introduction of the relative velocity information in Eq. (\ref{eqn.n5}) reduces the CA's invasiveness. However, due to the robot is controlled at the velocity level, the speed jump occurs in both Eq. (\ref{eqn.nn5}) and Eq. (\ref{eqn.n5}). Eq. (\ref{eqn.n4}) and our method ensure the smoothness of the CA trajectory. However, the performance index of our method is inferior than Eq. (\ref{eqn.n4}) in this example.

We further compare the performance of Eq. (\ref{eqn.n4}) and our method by gradually increasing the number of WMRs. In this paper, the enclosing radius of every WMR is valued as $0.48$, therefore, the accommodated maximum number of WMRs is $39$. The comparison results are shown in Table. \ref{tab2} and Fig. \ref{fig2}.
It is concluded from both Table. \ref{tab2} and Fig. \ref{fig2}, our method is competitive with Eq. (\ref{eqn.n4}), the better of this, our method still succeed in a crowed environment.
For Eq. (\ref{eqn.n4}), it provides minimal invasiveness. However, the introduction of the braking term slows down the WMRs' reaction speed, leading to collisions in crowed environments.

\subsubsection{Comparison of the deconfliction strategies}
Three commonly used deadlock detection mechanisms are:
﻿\begin{itemize}
\item[1)]The WMR's velocity is near-zero but its preferred velocity is non-zero \cite{Safety2017TRO}.
\item[2)]The WMR's position shows little change over a period of time.
\item[3)]The angle between the WMR's orientation and the line connecting the robot to an obstacle is near-zero \cite{he2024simultaneous}.
\end{itemize}
\begin{figure}
  \centering
  \setlength{\abovecaptionskip}{-0.01cm}   
 \includegraphics[width=3.45in]{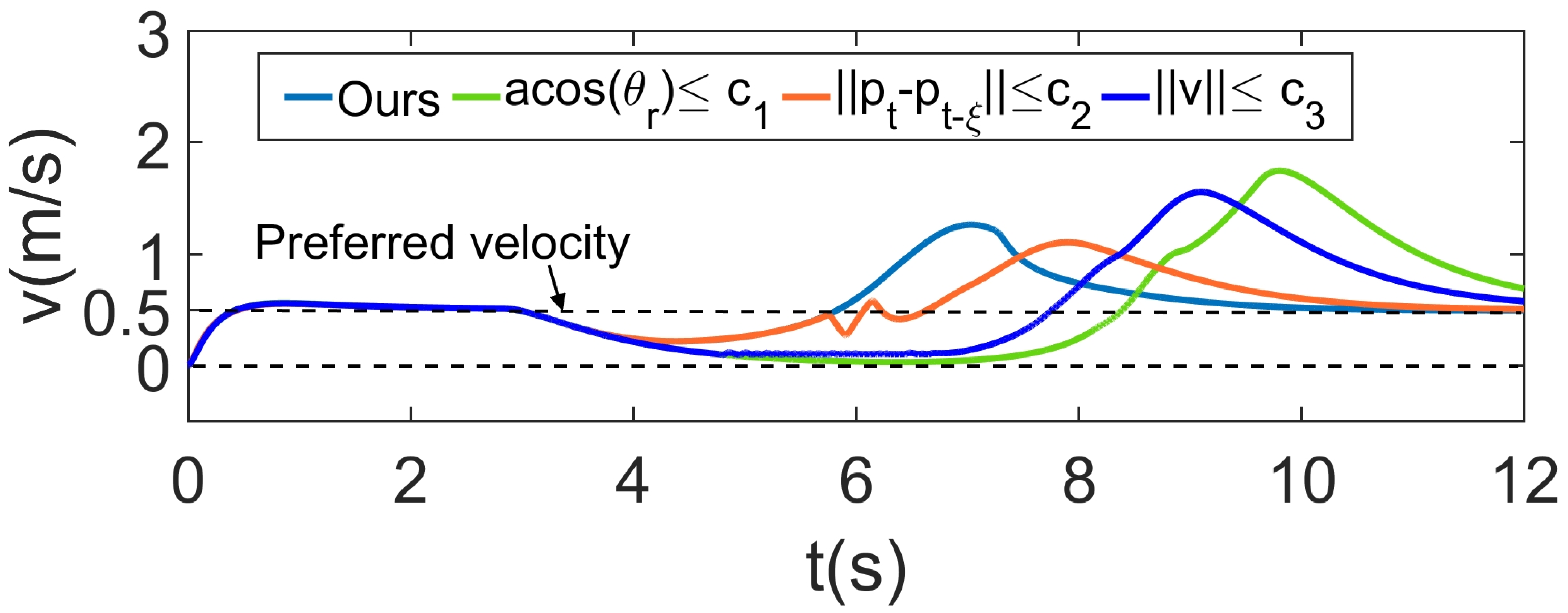}
  \caption{\justifying Velocity profiles synthesized by four deadlock detection methods, respectively, where deadlock decisions on near-zero velocity in \cite{Safety2017TRO,he2024simultaneous}, colored by the green and blue lines respectively, lead to a slow-reacting system.}\label{fig3a}
\end{figure}

 Fig. \ref{fig3a} gives velocity profiles synthesized by these three decision methods and our method. Obviously, both \cite{Safety2017TRO}
and  \cite{he2024simultaneous}, colored by the blue and green lines, lead to a slow-reacting system. Brick-red lines looks good, however, the determination of the dwell time $\xi$ makes it unable to adapt well to dynamic environments. Our method avoids the near-zero velocity and quickly returns to the desired position.

\begin{figure}[htb]
  \centering
\setlength{\abovecaptionskip}{-0.1cm}   
\begin{minipage}[]{5.5in}
\hspace{-10pt}
\subfigure[]{\includegraphics[scale=0.16]{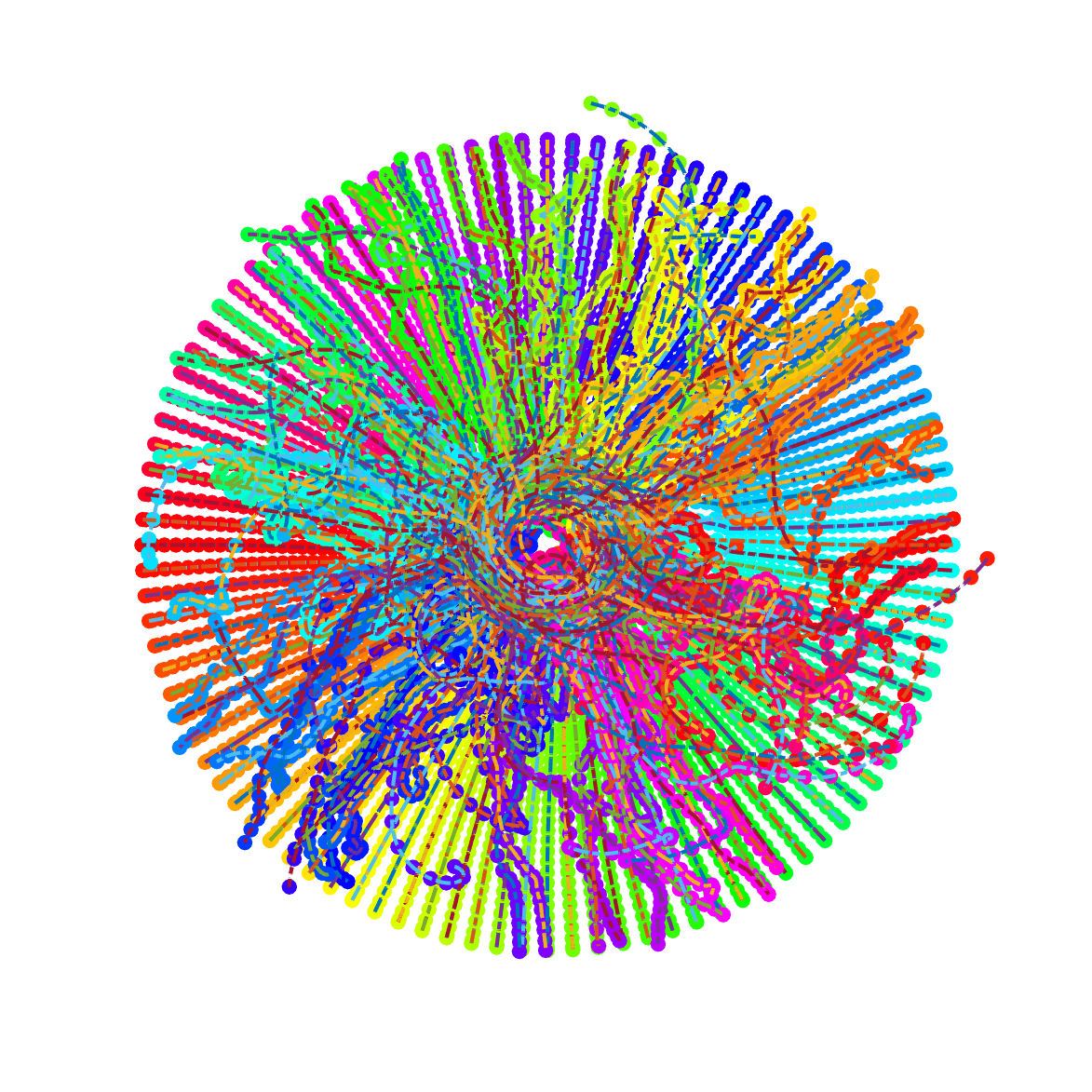}}
\hspace{-10pt}
\hspace{-10pt}
\subfigure[]{\includegraphics[scale=0.16]{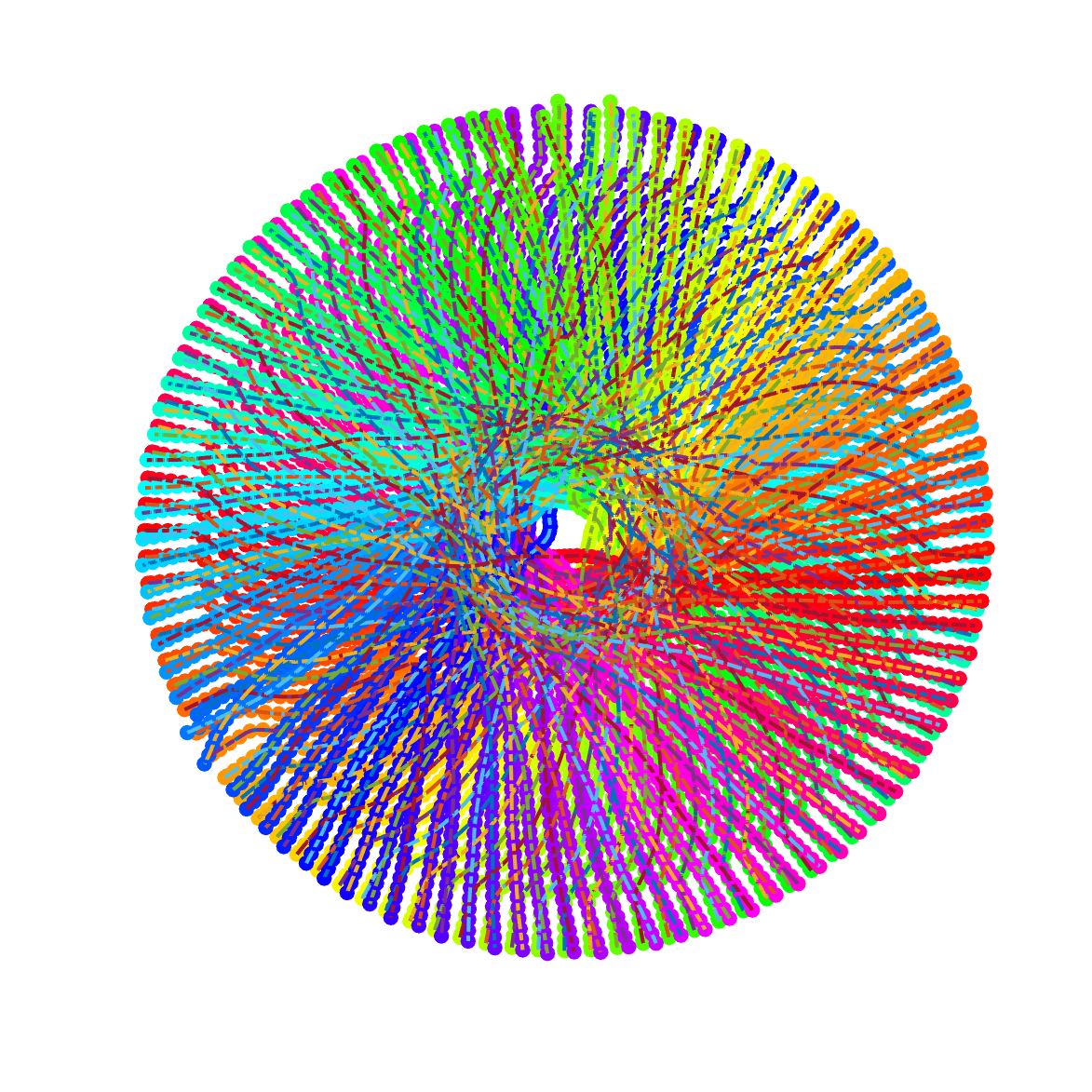}}
\hspace{-10pt}
\subfigure[]{\includegraphics[scale=0.16]{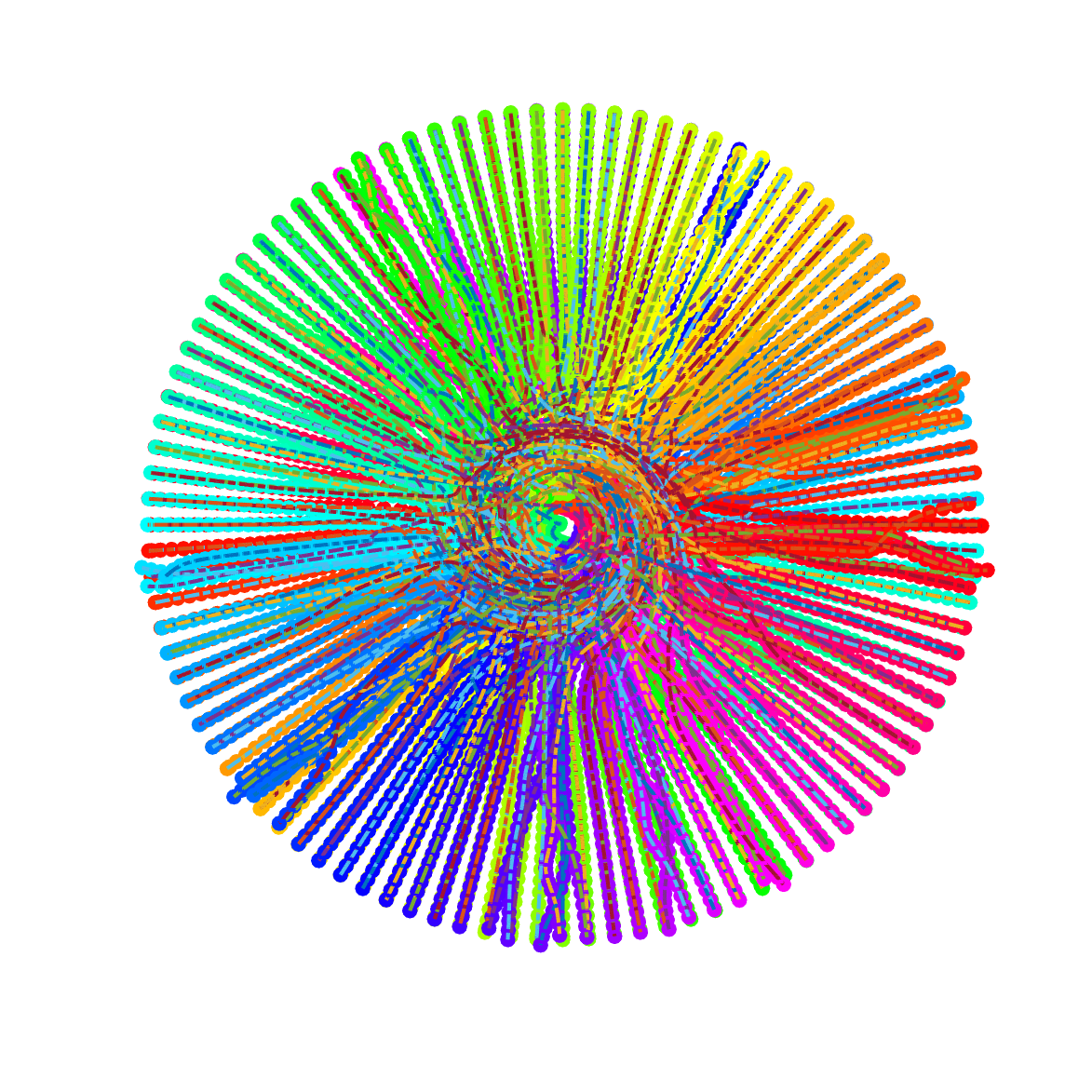}}
\end{minipage}
 \caption{\justifying 100 WMRs SOATT illustration achieved by: (a) Eq. (\ref{eqn.12a}), (b) Eq. (\ref{eqn.12b}), (c): Eq. (\ref{eqn.13}), respectively.}\label{fig.100}
 \end{figure}

Next, we conduct the comparison of the deconfliction strategies Eq. (\ref{eqn.12a}) \cite{Safety2017TRO}, Eq. (\ref{eqn.12c}) \cite{he2024simultaneous}, and our methods Eq. (\ref{eqn.12b}), Eq. (\ref{eqn.13}) on the CA's invasiveness. However, we do not give the visible and numerical results of Eq. (\ref{eqn.12c}) and the performance indexes due to this method is easy to lead to the Zeno phenomenon in our implemention. Moreover, the introduction of the virtual obstacle in \cite{he2024simultaneous} not only increases the computational account, but also a larger virtual space makes the CA be conservative.
 Table. \ref{tab3} gives comparison on CA's invasiveness between our two methods and Eq. (\ref{eqn.12a}). In this example, the total travel time is set as $20\mathrm{s}$ and the radius of the considered circle scenario is $10$.
 Fig. \ref{fig.100} gives an illustration of trajectory generated by these three methods where the number of WMRs is $100$.
Following Table. \ref{tab3} and Fig. \ref{fig.100}, it is observed that as the number of WMRs increases, the advantages of our method gradually become apparent. When considering 100 WMRs, Eq. (\ref{eqn.12a}) fails in returning their respective desired positions after avoiding collisions. Moreover, although Eq. (\ref{eqn.12b}) is effective, we say that attaching the auxiliary vector to the TT error, \emph{i.e.}, Eq. (\ref{eqn.13}), can derive the best performance, and has no influence on global convergence of TT error as proven in previous section.

\begin{table}
\centering
\setlength{\belowcaptionskip}{0.01cm}
\caption{Comparison of deconfliction strategies.} \label{tab3}
\newcommand{\minitab}[2][l]{\begin{tabular}{#1}#2\end{tabular}}
 \setlength{\tabcolsep}{0.6mm}{
\begin{tabular}{c|c|c|ccc}
  \hline
    \multirow{5}{*}{\makecell{Numbers of \\WMRs}} &\multirow{5}{*}{Methods}&\multicolumn{4}{c}{Metrics}\\
    \cline{3-6}
    &&\multirow{3}{*}{\makecell{Intervention\\ Time}}&\multicolumn{3}{c}{\makecell{Proximity on\\Desired Trajectory}}\\
    \cline{4-6}
    &&&RMSE & MAE & \makecell{Standard\\ Deviation}  \\
  \hline
    \multirow{3}{*}{20}
    &Eq. (\ref{eqn.12a}) & 6.9450 & 0.7805 & 0.4277 &0.6529  \\
      \cline{2-6}
    &Eq. (\ref{eqn.12b})& \textbf{7.1750} & \textbf{1.0364} & \textbf{0.6204}&\textbf{0.8302} \\
      \cline{2-6}
    &Eq. (\ref{eqn.13})& \textbf{7.4600} & \textbf{0.8699} & \textbf{0.4831} & \textbf{0.7234}\\
      \cline{2-6}
  \hline
     \multirow{3}{*}{30}
    &Eq. (\ref{eqn.12a}) & 11.1500 & 0.9713 & 0.6034 &0.7612  \\
      \cline{2-6}
    &Eq. (\ref{eqn.12b})& \textbf{10.1450} & \textbf{1.2187} & \textbf{0.8450}& \textbf{0.8782} \\
      \cline{2-6}
    &Eq. (\ref{eqn.13})& \textbf{10.0800} & \textbf{1.0462} & \textbf{0.6448} & \textbf{0.8238}\\
      \cline{2-6}
  \hline
    \multirow{3}{*}{40}
    &Eq. (\ref{eqn.12a}) & 14.9750 & 1.1155 & 0.7701 &0.8070  \\
      \cline{2-6}
    &Eq. (\ref{eqn.12b})& \textbf{13.0900} & \textbf{1.3643} & \textbf{1.0416}&\textbf{ 0.8812} \\
      \cline{2-6}
    &Eq. (\ref{eqn.13})& \textbf{13.0300} & \textbf{1.2061} & \textbf{0.8009} & \textbf{0.9017}\\
      \cline{2-6}
      \hline
       \multirow{3}{*}{50}
    &Eq. (\ref{eqn.12a}) & 18.6800 & 1.2439 & 0.9239 &0.8329  \\
      \cline{2-6}
    &Eq. (\ref{eqn.12b})& \textbf{16.3650} &\textbf{ 1.5311} & \textbf{1.2674}& \textbf{0.8591} \\
      \cline{2-6}
    &Eq. (\ref{eqn.13}) & \textbf{16.7950} & \textbf{1.3076} & \textbf{0.9172} & \textbf{0.9320}\\
      \cline{2-6}
      \hline
    \multirow{3}{*}{60}
    &Eq. (\ref{eqn.12a}) & 19.7400 & 1.4536 & 1.1483 &0.8914  \\
      \cline{2-6}
    &Eq. (\ref{eqn.12b})& \textbf{19.3350} & \textbf{1.8085} & \textbf{1.5748}& \textbf{0.8891} \\
      \cline{2-6}
    &Eq. (\ref{eqn.13})& \textbf{18.6150} & \textbf{1.4795} & \textbf{1.1080} & \textbf{0.9804}\\
      \cline{2-6}
 \hline
\end{tabular}}
\end{table}

 \begin{figure*}
  \centering
\setlength{\abovecaptionskip}{-0.1cm}   
\begin{minipage}[]{7.5in}
\subfigure[]{\includegraphics[scale=0.35]{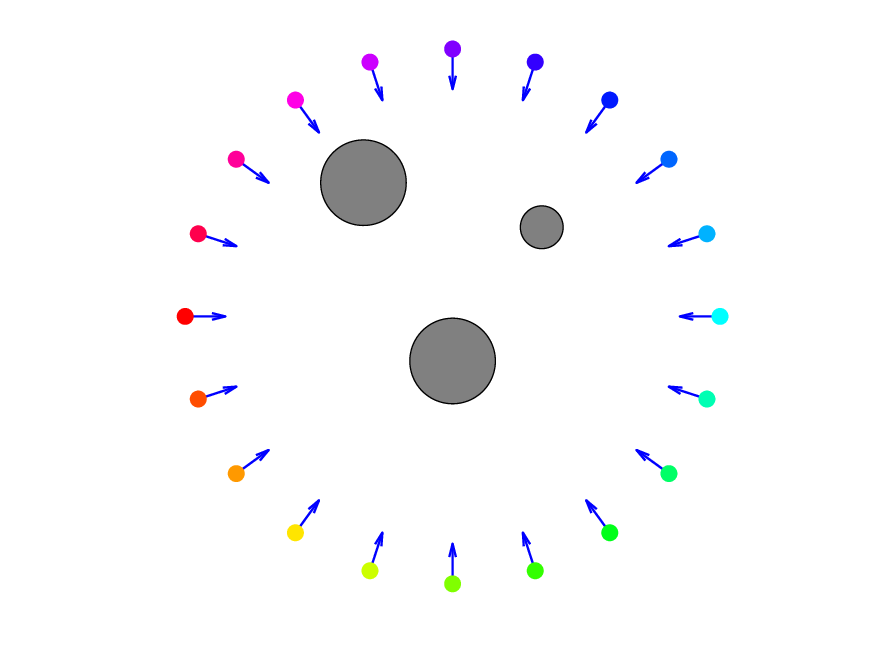}}
\hspace{-30pt}
\subfigure[]{\includegraphics[scale=0.35]{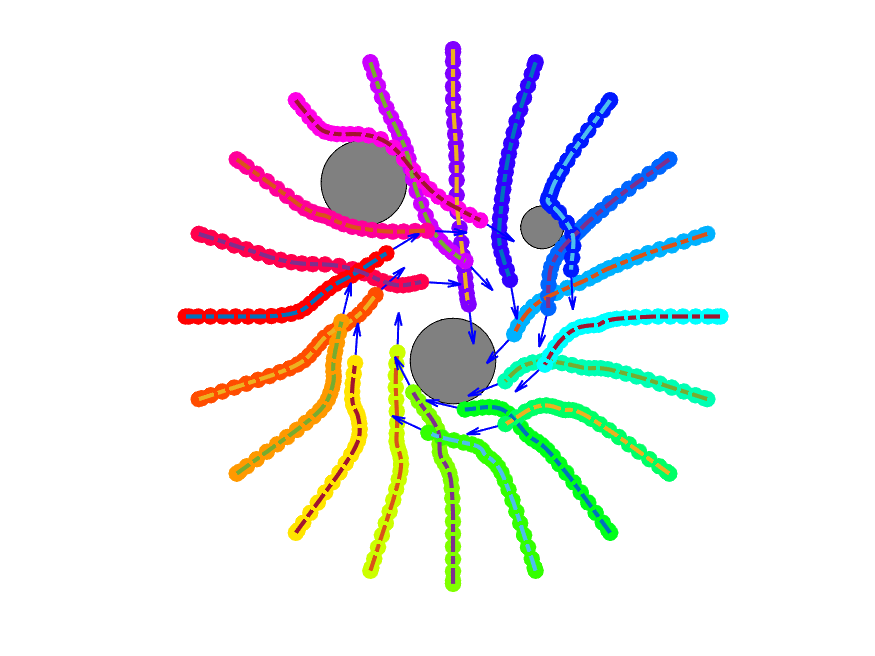}}
\hspace{-30pt}
\subfigure[]{\includegraphics[scale=0.35]{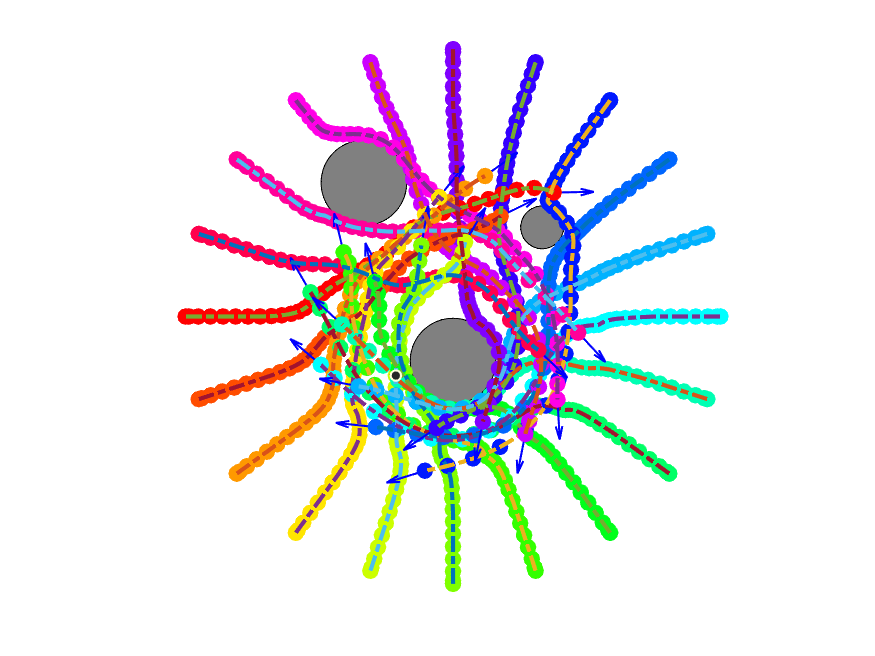}}
\hspace{-30pt}
\subfigure[]{\includegraphics[scale=0.35]{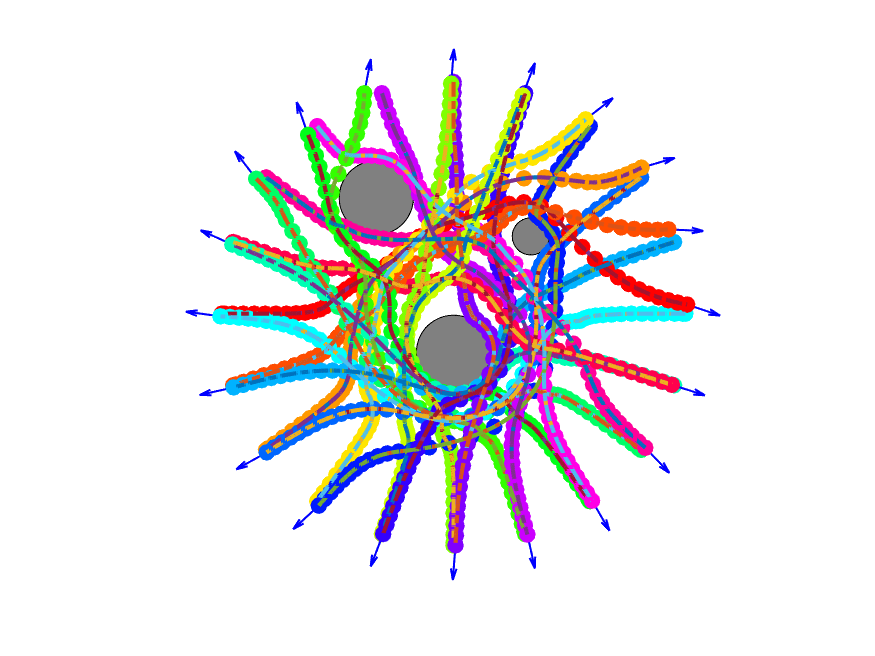}}
\end{minipage}
 \caption{Snapshot of twenty WMRs’ SOATT results obtained by the AA-SOATT method at different times, where three static obstacles are considered. (a) Initial states. (b) Collision avoidance. (c) WMRs return to their respective preferred trajectories. (d) Final states.}\label{fig5}
 \end{figure*}
\vspace{-10pt}
 \subsection{CA between MWMRs and Multi-obstacle}
In previous simulations, superiority of our method is shown by a series of comparison. In this part, we show feasibility of our method in a complex environment where WMRs and environmental obstacles coexist.
Considering this an example as shown in Fig. \ref{fig5}(a), twenty WMRs attempt to move through the center of a circle
to antipodal positions, where three static obstacles with different sizes are considered. Fig. \ref{fig5}(b)-Fig. \ref{fig5}(d) give snapshot of SOATT obtained by the proposed AA-SOATT method at different times. In this example, deconfliction strategy Eq. (\ref{eqn.13}) is employed. In Fig. \ref{fig5}(b), the CA module dominates the WMRs's behavior and as a result, WMRs deviate from their respective preferred trajectories to avoid collision with WMRs and obstacles. In Fig. \ref{fig5}(c), the TT module dominates the controller due to no potential collision risk is detected, WMRs start to return to their preferred trajectory until the destination is reached. Followed by Fig. \ref{fig5}(b)-Fig. \ref{fig5}(d), it is verified that our method is effective in a complicated environment.
\vspace{-10pt}
\subsection{Experiments verification}
\begin{figure}
  \centering
\setlength{\abovecaptionskip}{-0.1cm}   
\begin{minipage}[]{5in}
\subfigcapskip=-2pt
\subfigure[]{\includegraphics[scale=0.073]{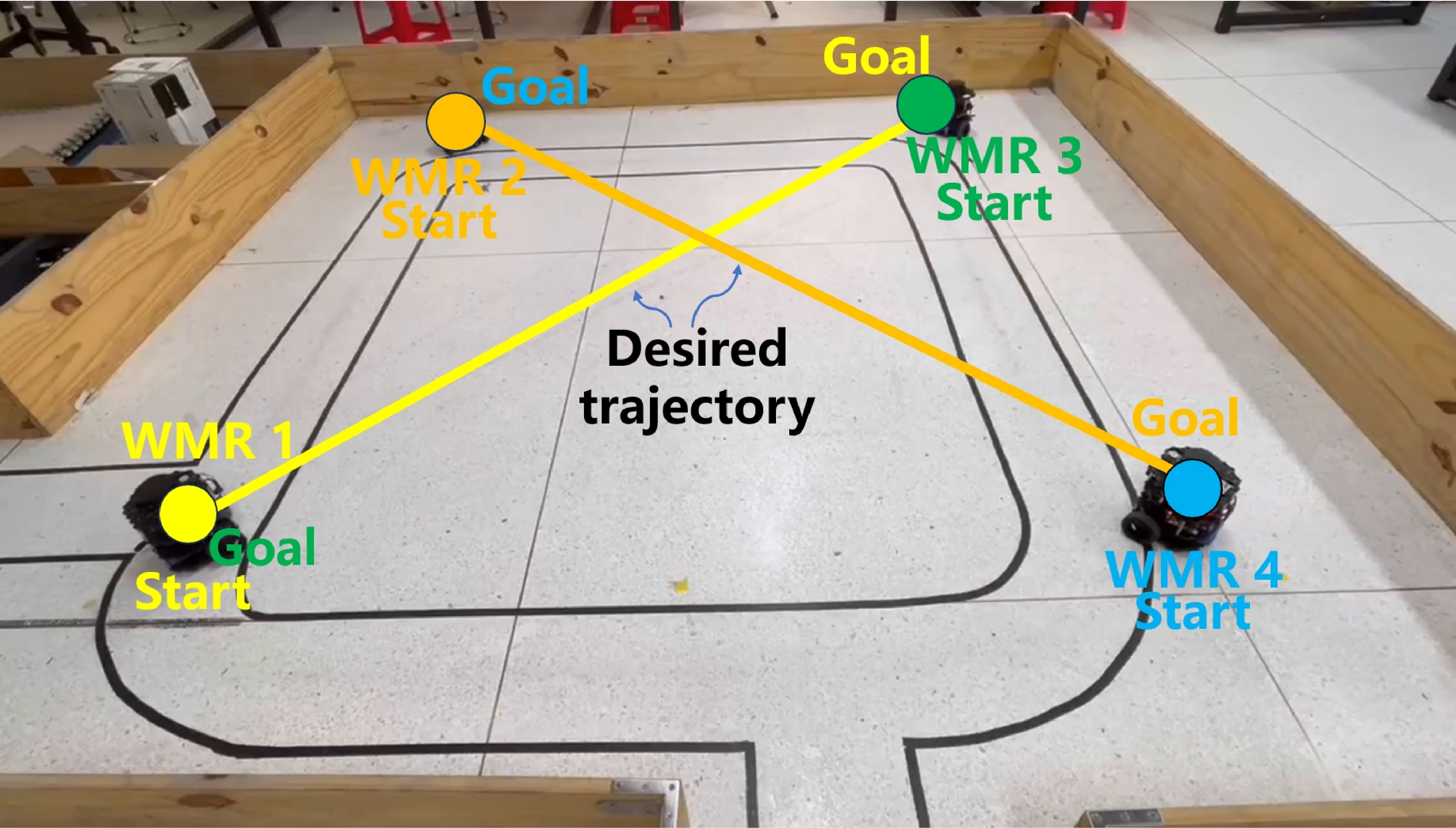}}
\hspace{3pt}
\subfigure[]{\includegraphics[scale=0.073]{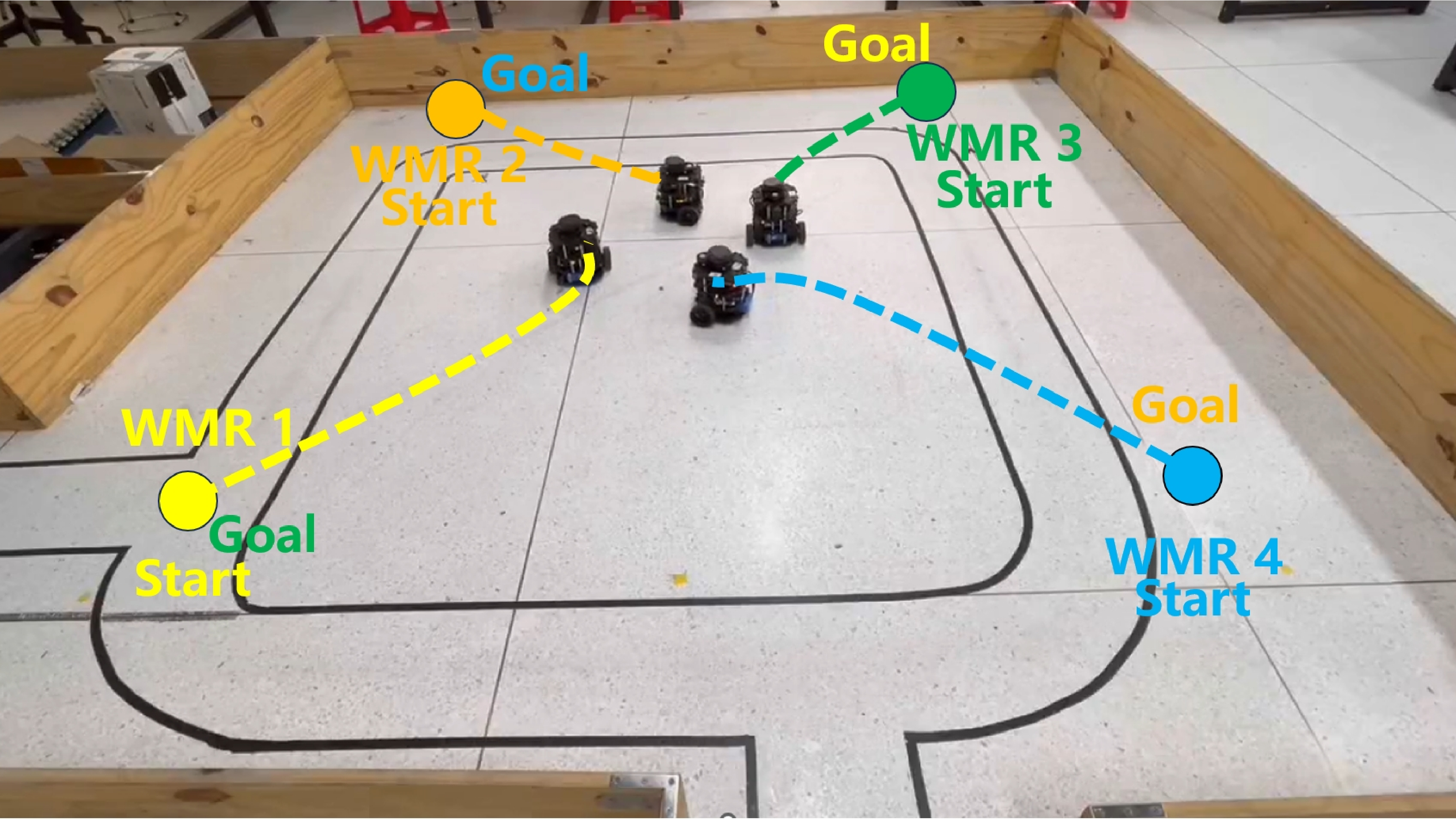}}
\end{minipage}
\begin{minipage}[]{5in}
\vspace{-2pt}
\subfigcapskip=-2pt
\subfigure[]{\includegraphics[scale=0.073]{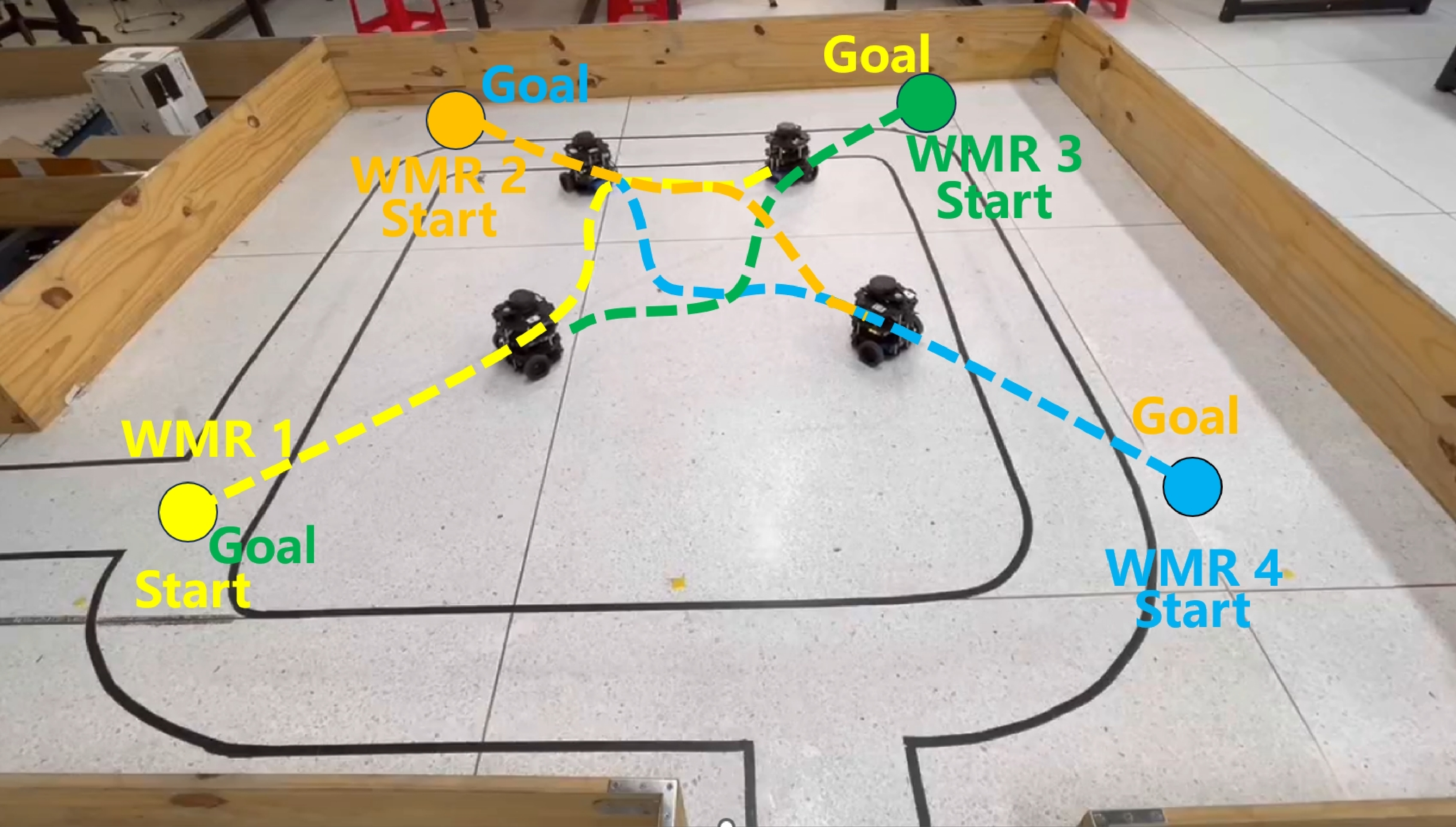}}
\hspace{3pt}
\subfigure[]{\includegraphics[scale=0.073]{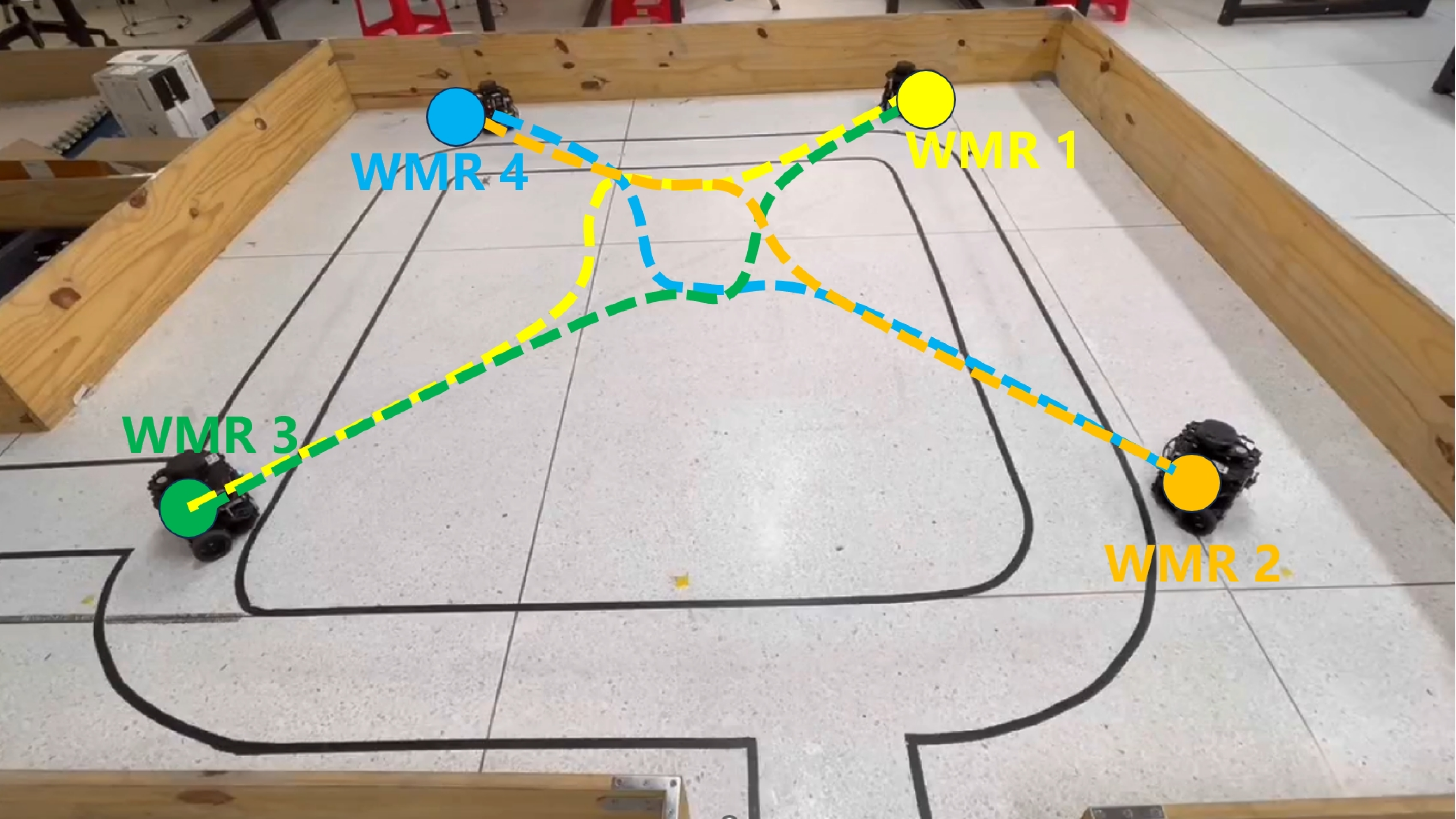}}
\end{minipage}
 \caption{\justifying MWMRs SOATT illustration. (a) Desired trajectory. (b) Collision avoidance behavior. (c) WMRs return to their respective desired trajectories when no potential collision risk. (d) Real trajectories achieved by the proposed method.}\label{fig.e3}
 \end{figure}

Based on the AA-SOATT method using Eq. (\ref{eqn.13}), physical experiment is implemented in TurtleBot 3 Burger. The obtained experimental results are shown in Fig. \ref{fig.e3}. In this experiment, we consider the position swapping scenario. As shown in Fig. \ref{fig.e3}(a), four WMRs, colored by four different colors,  are required to forward $2.24\mathrm{m}$ from their current positions at a speed of $0.1\mathrm{m/s}$. Both Fig. \ref{fig.e3}(b) and Fig. \ref{fig.e3}(c) show the collision avoidance snapshot, and Fig. \ref{fig.e3}(d) gives the final trajectories for every robot achieved by our proposed method. Following Fig. \ref{fig.e3}, it could be observed that every robot moves along its desired trajectory when no potential risk is detected. When the CA module is activated, every robot successfully avoids the impending collision by deviating from its preferred trajectory under the proposed CA strategy. WMRs return to their desired trajectories when the other three WMRs do not threaten its safety until its destination is reached.

\section{Conclusion}\label{sec.6}
An AA-SOATT method that is solved at the acceleration level has been proposed. It provides the solutions for a robot to avoid the impending collisions and deadlocks with its nearby WMRs while as close to its desired trajectory as possible. Compared to the existing studies, the constructed safe set can scale well with number of WMRs while maintains the CA path smoothing. The introduction of the auxiliary velocity vector in the error function has no influence on the convergence of TT. Extensive comparison indicate that our method can derive the minimal invasiveness when deployed in a large-scale robot system.
In the future, we will consider simultaneous obstacle avoidance and trajectory tracking for MWMRs systems in a narrow environment, \emph{e.g.}, corridor. Learning-and-prediction-based online MWMRs motion coordination is being considered deeply.

\bibliographystyle{IEEEtran}
\bibliography{reference}
\end{document}